\documentclass[10pt,journal,compsoc]{IEEEtran}

\usepackage[utf8]{inputenc} 
\usepackage[T1]{fontenc}    
\usepackage{hyperref}       
\usepackage{url}            
\usepackage{booktabs}       
\usepackage{amsfonts}       
\usepackage{nicefrac}       
\usepackage{microtype}      
\usepackage{xcolor}         
\usepackage{todonotes}

\usepackage{wrapfig}

\usepackage{colortbl}
\definecolor{myy}{RGB}{126,95,0}
\definecolor{mygray}{gray}{.85}
\definecolor{lightgray}{gray}{.95}
\definecolor{myblue}{RGB}{0, 0, 0}
\usepackage{multirow}
\usepackage{graphicx}
\newcommand{\tabincell}[2]{\begin{tabular}{@{}#1@{}}#2\end{tabular}}
\newcommand{\myparagraph}[1]{\vspace{1pt}\noindent{\bf #1}}
\usepackage{amsmath}

\usepackage{subfigure}
\usepackage{capt-of}
\usepackage{makecell}

\usepackage{mathtools}

\ifCLASSOPTIONcompsoc
  \usepackage[nocompress]{cite}
\else
  \usepackage{cite}
\fi

\ifCLASSINFOpdf
\else
\fi

\begin{document}
\title{
MTR++: Multi-Agent Motion Prediction with Symmetric Scene Modeling and Guided Intention Querying
}

\author{
Shaoshuai~Shi$^*$, 
Li~Jiang$^{*\dagger}$, 
Dengxin~Dai, 
and~Bernt~Schiele,~~\IEEEmembership{Fellow,~IEEE}

\IEEEcompsocitemizethanks{\IEEEcompsocthanksitem  Shaoshuai Shi, Li Jiang, Dengxin Dai, and Bernt Schiele are with the Max Planck Institute for Informatics, Saarland Informatics Campus, Germany.
\IEEEcompsocthanksitem {E-mail: \{sshi, lijiang, ddai, schiele\}@mpi-inf.mpg.de}
\IEEEcompsocthanksitem {$*$: equal contributions.}
\IEEEcompsocthanksitem {$\dagger$: corresponding author.}
}
}

\markboth{Journal of \LaTeX\ Class Files,~Vol.~14, No.~8, August~2015}%
{Shell \MakeLowercase{\textit{et al.}}: Bare Demo of IEEEtran.cls for Computer Society Journals}

\IEEEtitleabstractindextext{%
\begin{abstract}
Motion prediction is crucial for autonomous driving systems to understand complex driving scenarios and make informed decisions. However, this task is challenging due to the diverse behaviors of traffic participants and complex environmental contexts. In this paper, we propose Motion TRansformer (MTR) frameworks to address these challenges. The initial MTR framework utilizes a transformer encoder-decoder structure with learnable intention queries, enabling efficient and accurate prediction of future trajectories. By customizing intention queries for distinct motion modalities, MTR improves multimodal motion prediction while reducing reliance on dense goal candidates. The framework comprises two essential processes: global intention localization, identifying the agent's intent to enhance overall efficiency, and local movement refinement, adaptively refining predicted trajectories for improved accuracy. Moreover, we introduce an advanced MTR++ framework, extending the capability of MTR to simultaneously predict multimodal motion for multiple agents. MTR++ incorporates symmetric context modeling and mutually-guided intention querying modules to facilitate future behavior interaction among multiple agents, resulting in scene-compliant future trajectories. Extensive experimental results demonstrate that the MTR framework achieves state-of-the-art performance on the highly-competitive motion prediction benchmarks, while the MTR++ framework surpasses its precursor, exhibiting enhanced performance and efficiency in predicting accurate multimodal future trajectories for multiple agents. 
\end{abstract}

}

\maketitle

\IEEEdisplaynontitleabstractindextext

\IEEEpeerreviewmaketitle

\IEEEraisesectionheading{\section{Introduction}\label{sec:introduction}}

\IEEEPARstart{M}{otion} prediction constitutes a pivotal undertaking within the realm of contemporary autonomous driving systems, and it has gained significant attention in recent years due to its vital role in enabling robotic vehicles to understand driving scenarios and make judicious decisions \cite{gu2021densetnt, jia2021ide, tolstaya2021identifying, liu2021multimodal, ye2021tpcn, jia2022multi, ngiam2021scene,zhou2022hivt, jia2023towards}. 
The core of motion prediction lies in accurately anticipating the future actions of traffic participants by considering observed agent states and complex road maps. However, this task is challenging due to the inherent multimodal behaviors exhibited by agents and the intricacies of the surrounding environment.

\begin{figure}[t]
    \centering
    \includegraphics[width=0.99\linewidth]{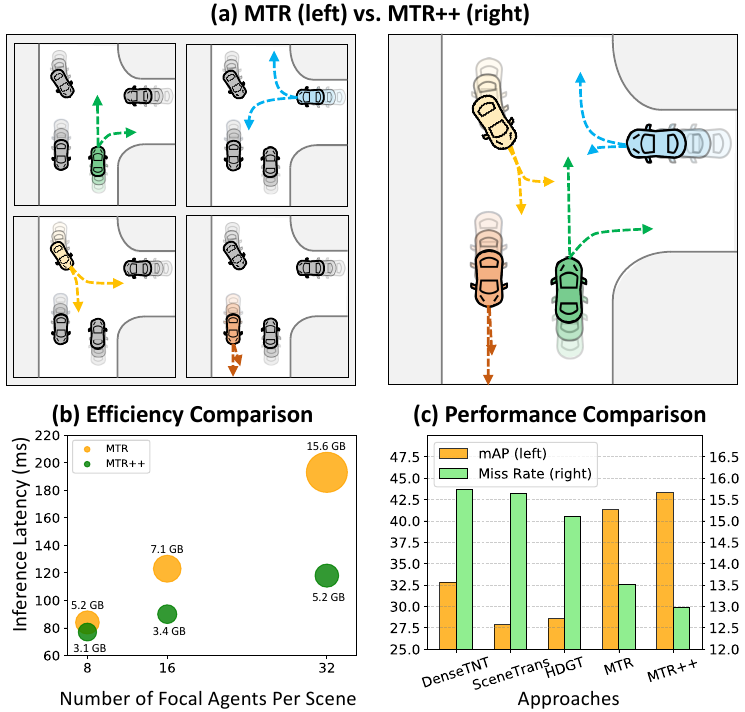}\\
    \vspace{-0.1in}
    \caption{The comparison of MTR and MTR++ frameworks.
    The MTR++ framework surpasses its predecessor, MTR, in several key aspects. In subfigure (a), MTR++ demonstrates its ability to predict the future trajectories of multiple agents simultaneously. Notably, in subfigure (b), MTR++ excels in both inference speed and memory efficiency, particularly when dealing with a larger number of interested agents. Additionally, as depicted in subfigure (c), the MTR++ framework outperforms both MTR and all other approaches, achieving superior performance overall.}
    \label{fig:teaser}
\end{figure}

To tackle these formidable challenges, prior studies~\cite{park2020diverse,liang2020learning,gao2020vectornet} have delved into diverse strategies aimed at encoding the complex scene context. Some works~\cite{ngiam2021scene,varadarajan2021multipath++} have employed the encoded agent features in motion decoders based on Multi-Layer Perceptrons (MLPs) to directly predict multiple potential future trajectories for the agent. Nonetheless, these methodologies generally exhibit a bias towards predicting the most frequently occurring modes observed in the training data, thereby yielding homogeneous trajectories that inadequately capture the agent's multimodal behaviors.
To improve trajectory predictions encompassing all potential future behaviors for the agent, alternative approaches~\cite{zhao2020tnt,gu2021densetnt} have explored a goal-based strategy. This strategy involves using a dense set of goal candidates representing feasible destinations for the agent. By predicting the probability associated with each candidate being the actual destination, these methods generate a full trajectory for each selected goal candidate. While this strategy reduces trajectory uncertainty during model optimization, the performance of such methods is highly dependent on the density of goal candidates. Fewer candidates lead to decreased performance, while an excessive number of candidates significantly increases computation and memory costs.
To address these challenges and enhance multimodal motion prediction while reducing reliance on dense goal candidates, we propose a novel collection of frameworks called Motion TRansformer (MTR) frameworks. These frameworks consist of an initial MTR framework and an advanced MTR++ framework.

In the MTR frameworks, we introduce a novel set of learnable intention queries integrated into a transformer encoder-decoder structure, which facilitates efficient motion prediction by employing each intention query to encompass the behavior prediction of a bunch of potential trajectories directed towards the same region. Guided by these intention queries, the MTR frameworks optimize two key tasks simultaneously. 
The first task is global intention localization, which aims to  roughly identify the agent's intention, thereby enhancing overall efficiency. 
The second task is local movement refinement, which strives to adaptively refine the predicted trajectory for each intention, thereby improving accuracy. The proposed MTR frameworks not only foster a stable training process without depending on dense goal candidates but also enable flexible and adaptable prediction by facilitating local refinement for each motion mode. 

Specifically, the MTR frameworks introduce distinct learnable intention queries to handle trajectory prediction across different motion modes. To accomplish this, a limited number of spatially distributed intention points (e.g., 64 in our case) are initially generated for each category. These intention points effectively reduce uncertainty in future trajectories by encompassing both motion direction and velocity.  
Each intention query represents the learnable position embedding of a specific intention point, assuming responsibility for predicting the future trajectory of that particular motion mode. This approach not only enhances multimodal motion prediction by explicitly leveraging different queries for different modes but also eliminates the necessity of dense goal candidates, as each query assumes responsibility for a substantial destination region. Moreover, the MTR frameworks employ the classification probability of all intention queries to roughly localize the agent's motion intention, while the predicted trajectory of each intention query undergoes iterative refinement through stacked transformer decoder layers. This iterative refinement process involves continually retrieving fine-grained local features of each trajectory. Our experiments show that these two complementary processes have demonstrated remarkable efficacy in predicting multimodal future motion.

In contrast to the initial MTR framework presented in our previous version~\cite{shi2023mtr}, which focuses on the multimodal motion prediction of a single agent, we introduce an advanced MTR++ framework that extends the capability to predict multimodal motion concurrently for multiple agents (see Fig.~\ref{fig:teaser}). Instead of individually encoding the scene context around each agent as in the previous version, we propose a novel symmetric scene context modeling strategy. This strategy employs a shared context encoder to symmetrically encode the entire scene for each agent,  incorporating a novel query-centric self-attention module to jointly capture the intricate scene context information within their respective local coordinate systems.
Furthermore, we introduce mutually-guided intention quering module in the motion decoder network, enabling agents to interact and influence each other's behavior. This facilitates more precise and scene-compliant joint motion prediction for multiple agents. Through these two enhancements, experimental results demonstrate that compared to the initial  MTR framework, the MTR++ framework effectively predicts more accurate multimodal future trajectories for multiple agents simultaneously. Additionally, as shown in Fig.~\ref{fig:teaser}, the efficiency advantage of the MTR++ framework becomes more pronounced as the number of agents increases.

Our contributions are four-fold: 
(1) We introduce the MTR frameworks, which incorporate a novel set of learnable intention queries within the transformer encoder-decoder architecture for motion prediction. By customizing intention queries to address distinct motion modalities, the MTR frameworks not only achieve more precise multimodal future trajectory predictions that encompass a wide range of possibilities but also obviate reliance on dense goal candidates. 
(2) We propose the advanced MTR++ framework for simultaneous multimodal motion prediction of multiple agents. This framework incorporates two key components: a symmetric scene context modeling module that allows for shared context encoding among multiple agents, and a mutually-guided intention querying module that facilitates the interaction of agents' future behaviors and enables the prediction of scene-compliant trajectories.
(3) The initial MTR framework achieves state-of-the-art performance on the motion prediction benchmark of Waymo Open Motion Dataset (WOMD)~\cite{ettinger2021large}, surpassing previous ensemble-free approaches with a remarkable mAP gain of +$8.48\%$. Additionally, the MTR++ framework further enhances the capabilities of the initial MTR framework, enabling concurrent joint multimodal motion prediction for multiple agents and improving both performance and efficiency. 
(4) Notably, our initial MTR and MTR++ frameworks won the championship of the highly-competitive Waymo Motion Prediction Challenge in 2022\cite{womd_leaderboard2022} and 2023~\cite{womd_leaderboard2022}, respectively, demonstrating their superiority and effectiveness.

\section{Related Work}

\myparagraph{Scene Context Encoding for Motion Prediction.}
The motion prediction task in autonomous driving scenarios involves the encoding of the input road map and agent history states to generate future trajectories of the agent, which plays a crucial role in this task. Prior works~\cite{park2020diverse,marchetti2020mantra,casas2020spagnn,djuric2020uncertainty,zhang2020novel,biktairov2020prank,casas2021mp3} have commonly employed rasterization techniques to convert the scene context into images, allowing for processing with convolutional neural networks (CNNs). LaneGCN~\cite{liang2020learning} utilizes a lane graph to capture the topological information of the map, and recent works~\cite{gu2021densetnt,sun2022m2i,ngiam2021scene,varadarajan2021multipath++} have widely adopted VectorNet~\cite{gao2020vectornet}  for its efficiency and scalability. VectorNet represents both road maps and agent trajectories as polylines. In our MTR frameworks, we also adopt this vector representation. However, instead of constructing a global graph of polylines, we advocate employing a transformer encoder on a locally connected graph. This strategy not only better preserves the input's locality structure but also improves memory efficiency, enabling larger map encodings for long-term motion prediction.

\myparagraph{Multimodal Future Behavior Modeling.} Given the encoded scene context features, existing works explore diverse strategies for modeling the agent's multimodal future behaviors. Early works~\cite{alahi2016social,gupta2018social,rhinehart2018r2p2,tang2019multiple,rhinehart2019precog} suggests generating a set of trajectory samples to approximate the output distribution. other studies~\cite{chai2019multipath,hong2019rules,mercat2020multi,phan2020covernet,salzmann2020trajectron++} parameterize multimodal predictions with Gaussian Mixture Models (GMMs) to generate compact distribution. The HOME series~\cite{gilles2021home,gilles2021gohome} generates trajectories by sampling a predicted heatmap. IntentNet~\cite{casas2018intentnet} considers intention prediction as a classification problem involving eight high-level actions, while \cite{liu2021multimodal} proposes a region-based training strategy.
Goal-based methods~\cite{zhao2020tnt,rhinehart2019precog,fang2020tpnet,mangalam2020not} represent another category, estimating several agent goal points before completing the full trajectory for each goal. 

The large-scale Waymo Open Motion Dataset (WOMD)\cite{ettinger2021large} has recently been introduced for long-term motion prediction. To address this challenge, DenseTNT~\cite{gu2021densetnt} employs a goal-based strategy to classify trajectory endpoints from dense goal candidates. Other works directly predict the future trajectories based on the encoded agent features~\cite{ngiam2021scene} or latent anchor embedding~\cite{varadarajan2021multipath++}. 
Nonetheless, the goal-based strategy raises efficiency concerns due to the numerous goal candidates, while the direct-regression strategy converges slowly and exhibits a bias to predict homogeneous trajectories since various motion modes are regressed from identical agent features. In contrast, our MTR frameworks employ a small set of learnable intention queries to address these limitations, facilitating the generation of future trajectories with extensive modalities and eliminating numerous goal candidates by employing mode-specific learnable intention queries to predict different motion modes.

Furthermore, we employ Gaussian Mixture Models in tandem with intention queries to model the continuous distribution of an agent's multimodal future behavior at each time step. This approach yields a parametric multimodal future distribution as the output, capable of generating occurrence probabilities for any specified future trajectory. In contrast, utilizing a set of sparse trajectories fails to provide a continuous and compact future distribution. Similarly, predicting dense future heatmaps incurs a substantial computational cost, necessitating a compromise between resolution and computational efficiency.

\myparagraph{Simultaneous Motion Prediction of Multiple Agents.}
In predicting an individual agent's future trajectories, state-of-the-art works~\cite{gu2021densetnt,varadarajan2021multipath++}, including our previous version~\cite{shi2023mtr}, typically customize the scene context encoding for that agent by normalizing all inputs centered on it. This strategy results in computational inefficiencies when predicting motion for multiple agents. To simultaneously predict future trajectories for multiple agents, SceneTransformer~\cite{ngiam2021scene} encodes all road graphs and agents into a scene-centric embedding applicable to all agents. However, their feature encoding still relies on a global coordinate system centered on an agent of interest (\emph{e.g.}, the autonomous vehicle), limiting its performance for off-center agents. 
Recent works~\cite{zhou2022hivt,jia2022hdgt} explore encoding the agents' node features in an ego-centric coordinate system, while they generally construct hand-crafted relation graphs and alternate node-edge updating strategy. 
In contrast, our MTR++ framework introduces symmetric scene context modeling for all agents with innovative query-centric self-attention, operating on a straightforward polyline graph using the native transformer encoder module with relative position encoding, thereby promoting more efficient and concise shared scene context encoding.

To enable the behavioral interaction of multiple agents, recent research M2I~\cite{sun2022m2i} introduces a triad of models, initially employing a relation predictor to categorize two interacting agents as influencer and reactor, followed by the sequential generation of their future trajectories via a marginal predictor and a conditional predictor, respectively. 
Conversely, our MTR++ framework integrates mutually-guided intention queries, fostering the behavioral interaction of more than two agents within a unified model, wherein their predicted future behaviors naturally interact through stacked transformer decoder layers, thereby yielding superior scene-compliant trajectories with higher efficiency for multiple agents.

\myparagraph{Transformer.}
Transformer~\cite{vaswani2017attention} has been extensively employed in natural language processing~\cite{devlin2018bert,bao2021beit} and computer vision~\cite{dosovitskiy2021an,wang2018non,carion2020end,wang2021multi,zeng2021motr,wang2023dsvt}. 
Our approach draws inspiration from DETR~\cite{carion2020end} and its subsequent works~\cite{zhu2020deformable,meng2021conditional,yang2022unified,lai2022stratified,liu2022dabdetr,chen2022mppnet,zhang2022dino}, particularly DAB-DETR~\cite{liu2022dabdetr}, where the object query serves as the positional embedding of an anchor box. 
Motivated by their notable success in object detection, we introduce the innovative concept of learnable intention query to model multimodal motion prediction with prior intention points. Each intention query is tasked with predicting a specific motion mode and enables iterative motion refinement by integrating with transformer decoders.

\begin{figure*}[t]
    \centering
    \includegraphics[width=0.99\linewidth]{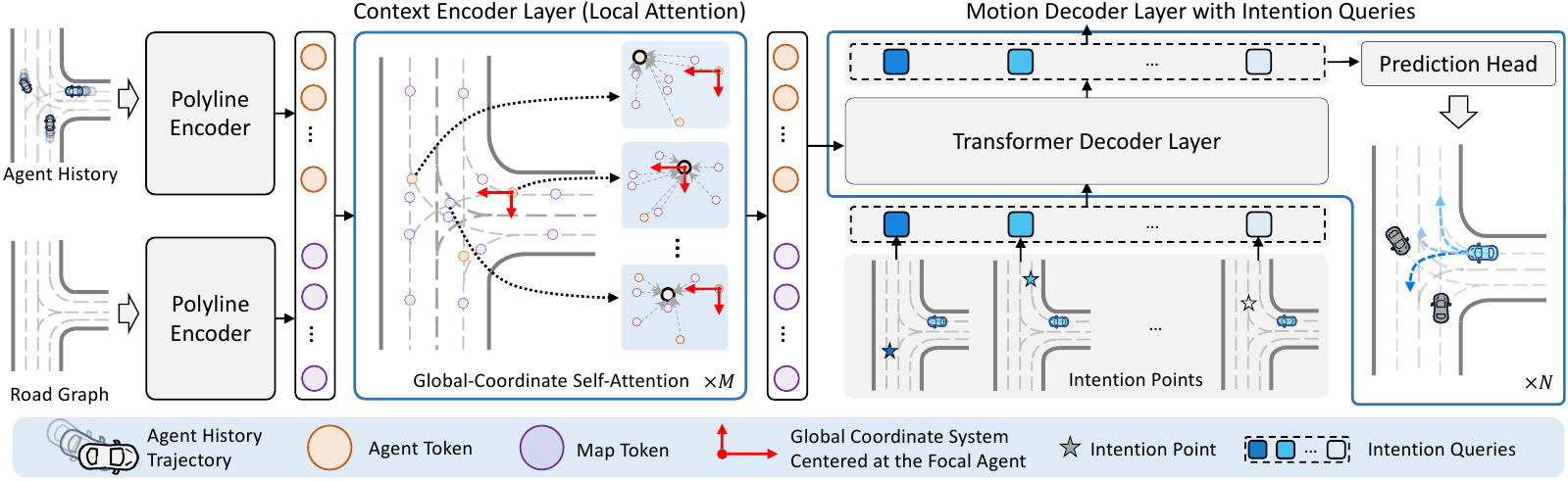}\\
    \vspace{-0.1in}
    \caption{
    The architecture of the MTR framework. In this framework, we first utilize two polyline encoders to encode the polylines derived from agent history trajectories and road lanes into token features. Next, multiple transformer encoder layers with local self-attention are utilized to model the relationships among different tokens within the global coordinate system centered around the focal agent of interest. This allows for a comprehensive understanding of the scene contextual information. Finally, a small set of learnable intention queries are integrated into the stacked transformer decoder layers to aggregate information from the encoded context features. Each intention query is responsible for predicting future trajectories towards a specific intention point, enabling the generation of multimodal future trajectories for the focal agent.
}
    \label{fig:mtr_framework}
\vspace{-3mm}
\end{figure*}

\section{MTR for Multimodal Motion Prediction}
We propose Motion TRansformer (MTR), which adopts a novel transformer encoder-decoder architecture incorporating iterative motion refinement for predicting multimodal future motion.
The overall framework is presented in Fig.~\ref{fig:mtr_framework}. 
In Sec.~\ref{sec:mtr_encoder}, we introduce our encoder network for scene context modeling. 
In Sec.~\ref{sec:mtr_decoder}, we present motion decoder network with a novel concept of intention query for predicting multimodal trajectories. 
Finally, in Sec.~\ref{sec:mtr_gmm}, we introduce the optimization process of our framework.

\subsection{Transformer Encoder for Scene Context Modeling}\label{sec:mtr_encoder}
The forthcoming actions of the agents are greatly influenced by their interactions and the road map. To incorporate this contextual information into the model, prior approaches have employed diverse techniques, such as constructing a comprehensive interactive graph~\cite{gao2020vectornet,gu2021densetnt} or condensing map features into agent-specific features~\cite{ngiam2021scene,varadarajan2021multipath++}. However, we argue that preserving the locality structure of the scene context, particularly the road map, is crucial. Thus, we introduce a transformer encoder network that utilizes local self-attention to better capture this structural information.

\myparagraph{Input Representation with Single Focal Agent.} 
We adopt the vectorized depiction~\cite{gao2020vectornet} to arrange both input trajectories and road maps as polylines.  When forecasting the motion of an individual focal agent, we employ the \textit{focal-agent-centric} approach~\cite{zhao2020tnt,gu2021densetnt,varadarajan2021multipath++}, which normalizes all inputs to the global coordinate system centered on this agent.

Concretely, the past states of $N_a$ agents are denoted as $S_{\text{A}}^{\text{(g)}}\in \mathbb{R}^{N_a\times T_h \times C_a}$ {(where ``g'' indicating the global reference frame)}. Here, $T_h$ represents the duration of the historical observations, and $C_a$ corresponds to the dimensionality of the state information, encompassing factors such as position, orientation, and velocity.
Zero-padding is applied to the positions of absent frames in trajectories comprising fewer than $T_h$ frames. 
The road map is represented as $S_{\text{M}}^{\text{(g)}}\in \mathbb{R}^{N_m\times n \times C_m}$, where $N_m$ indicates the number of map polylines, $n$ represents the number of points in each polyline, and $C_m$ signifies the number of attributes for each point (e.g., location and road type). 
Both $S_{\text{A}}^{\text{(g)}}$ and $S_{\text{M}}^{\text{(g)}}$ are encoded utilizing a PointNet-like~\cite{qi2017pointnet} polyline encoder as:
\begin{align}\label{eq:mtr_polyline}
F_{\text{A}}^{\text{(g)}} = \phi\left(\text{MLP}(S_{\text{A}}^{\text{(g)}})\right),~~~~ F_{\text{M}}^{\text{(g)}} = \phi\left(\text{MLP}(S_{\text{M}}^{\text{(g)}})\right),
\end{align}
where $\text{MLP}(\cdot)$ represents a multi-layer perceptron, while $\phi$ denotes max-pooling, employed to encapsulate each polyline's feature as agent features $F_{\text{A}}^{\text{(g)}}\in \mathbb{R}^{N_a\times D}$ and map features $F_{\text{M}}^{\text{(g)}} \in \mathbb{R}^{N_m\times D}$ with a feature dimension of $D$.

These two types of polyline features are concatenated to form the following input token features, denoted as 
$F_{\text{AM}}^{\text{(g)}}=[F_{\text{A}}^{\text{(g)}}, F_{\text{M}}^{\text{(g)}}] \in \mathbb{R}^{(N_a+N_m)\times D}$. 
The positions of these tokens are denoted as 
$P_{\text{AM}}^{\text{(g)}}=[P_\text{A}^{\text{(g)}}, P_\text{M}^{\text{(g)}}] \in \mathbb{R}^{(N_a+N_m)\times 2}$, where we utilize the most recent positions for agent tokens (denoted as $P_\text{A}^{\text{(g)}} \in \mathbb{R}^{N_a\times 2}$) and polyline centers for map tokens (denoted as $P_\text{M}^{\text{(g)}} \in \mathbb{R}^{N_m\times 2}$).

\myparagraph{Scene Context Encoding with Local Transformer Encoder.}
The local structure of scene context is vital for motion prediction. For instance, the relationship between two parallel lanes is essential for modeling lane-changing behavior, but utilizing attention on a globally connected graph treats all lane relations equally. Therefore, we incorporate prior knowledge into the context encoder by employing local attention, which better preserves the locality structure and is more memory-efficient. Specifically, the attention module of each transformer encoder layer can be expressed as:
\begin{align}\label{eq:mtr_encoder_sa}
\begin{split}
{{F'}^{\text{(g)}}_{\text{AM}}}{\scriptstyle [i]}=\text{MHSA}\bigl(&\text{Q:~}[F_{\text{AM}}^{\text{(g)}}{\scriptstyle [i]}, \text{PE}{(P_{\text{AM}}^{\text{(g)}}{\scriptstyle [i]})]}, \\
&\text{K:~}\{[F_{\text{AM}}^{\text{(g)}}{\scriptstyle [j]}, \text{PE}{(P_{\text{AM}}^{\text{(g)}}{\scriptstyle [j]})}]\}_{j\in \Omega(i)},\\ 
&\text{V:~}\{F_{\text{AM}}^{\text{(g)}}{\scriptstyle [j]}\}_{j\in \Omega(i)}\bigr), 
\end{split}
\end{align}
where $i \in \{1, \dots, N_a+N_m\}$. $\Omega(i)$ indicates the index set of the $k$ neighborhoods of $i$-th token. $\text{MHSA}(\cdot_\text{query}, ~\cdot_\text{key}, ~\cdot_\text{value})$ denotes multi-head self-attention layer~\cite{vaswani2017attention}.  $\text{PE}(\cdot)$ signifies the sinusoidal positional encoding of input tokens. 
${{F'}^{\text{(g)}}_{\text{AM}}}{\scriptstyle [i]} \in \mathbb{R}^D$ is the output feature of the $i$-th token of this encoder layer.
Thanks to this local self-attention, our framework can encode a considerably larger scene context.

By stacking multiple transformer encoder layers, the encoder network generates the token features ${{F'}^{\text{(g)}}_{\text{AM}}}\in \mathbb{R}^{(N_a+N_m)\times 2}$. We decompose these features to obtain the agent history features ${F^{\text{(g, past)}}_{\text{A}}}\in \mathbb{R}^{N_a\times D}$ and map features $F^{\text{(g)}}_{\text{M}} \in \mathbb{R}^{N_m\times D}$, where the agent history features will be further enhanced as ${F^{\text{(g)}}_{\text{A}}}\in \mathbb{R}^{N_a\times D}$ by the following dense future prediction module. Note that in the following sections, we employ the same notations for convenience, referring to ${F^{\text{(g)}}_{\text{A}}}\in \mathbb{R}^{N_a\times D}$ and $F^{\text{(g)}}_{\text{M}} \in \mathbb{R}^{N_m\times D}$ to represent the agent features and map features, respectively, which are encoded by the context encoder.

\myparagraph{Dense Future Prediction for All Agents.}
Interactions with other agents significantly influence the behaviors of our focal agent. 
Existing methods, such as hub-host networks~\cite{zhu2019starnet}, dynamic relational reasoning~\cite{li2020evolvegraph}, and social spatial-temporal networks~\cite{xu2021tra2tra}, mainly focus on learning past interactions but often overlook future trajectory interactions. 
To compensate for this limitation, we propose a method that densely predicts future states for all agents using a straightforward regression head on the encoded history features $F_{\text{A}}^{\text{(g, past)}}$, as follows
\begin{align}\label{eq:aux_reg}
S_{\text{A}}^{\text{(g, future)}}=\text{MLP}(F_{\text{A}}^{\text{(g, past)}}),
\end{align}
where $S_{\text{A}}^{\text{(g, future)}} \in \mathbb{R}^{N_a\times (T_f \times 4)}$ includes the future position and velocity of each agent, and $T_f$ denotes the number of future frames to be predicted. The predicted trajectories $S_{\text{A}}^{\text{(g, future)}}$ are encoded using the same polyline encoder as in Eq.~\eqref{eq:mtr_polyline}, producing features $F_{\text{A}}^{\text{(g, future)}}\in \mathbb{R}^{N_a\times D}$. These features are combined with $F_{\text{A}}^{\text{(g, past)}}\in \mathbb{R}^{N_a\times D}$ using feature concatenation and three MLP layers, resulting in enhanced features $F_{\text{A}}^{\text{(g)}}\in \mathbb{R}^{N_a\times D}$. 

By supplying the motion decoder network with additional future context information, this approach effectively improves the model's capability to predict more accurate future trajectories for the focal agent. Experimental results demonstrate that this simple auxiliary task effectively enhances the performance of multimodal motion prediction.

\subsection{Motion Decoder with Intention Query} \label{sec:mtr_decoder}
To facilitate multimodal motion prediction, the MTR framework utilizes a transformer-based motion decoder network that incorporates the previously encoded scene context features. We introduce the concept of intention query, which facilitates multimodal motion prediction through the joint optimization of global intention localization and local movement refinement. 
As depicted in Fig.~\ref{fig:mtr_framework}, the motion decoder network consists of stacked transformer decoder layers that iteratively refine predicted trajectories utilizing learnable intention queries. Next, we elaborate on the detailed structure.

\begin{figure}[t]
	\centering
	\includegraphics[width=0.99\linewidth]{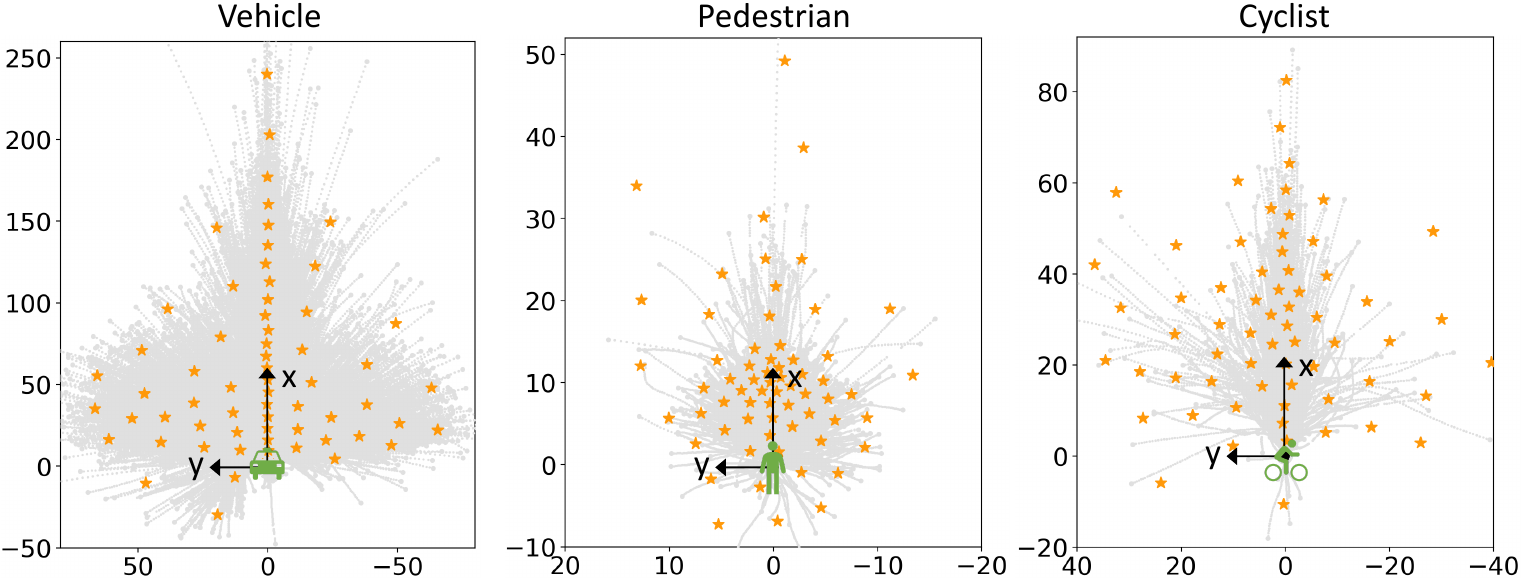}\\
	\vspace{-2mm}
	\caption{The distribution of intention points for each category, where the intention points are shown as orange stars. The gray dotted lines indicate the distribution of ground-truth trajectories for each category, and note that only 10\% ground-truth trajectories in the training dataset are drawn in the figure for better visualization.}
	\label{fig:cluster}
	\vspace{-3mm}
\end{figure}

\myparagraph{Learnable Intention Query\footnote{
To streamline the illustration of the motion decoder, we simplify the two components of the motion query pair in our previous version~\cite{shi2023mtr} by using the new concept of intention query.}.}
To efficiently and precisely pinpoint an agent's potential motion intentions, we propose the learnable \emph{intention query} to diminish the uncertainty of future trajectories by employing different intention queries for different motion modes. Specifically, for each category, we generate $\mathcal{K}$ representative intention points $I^{(\text{s})} \in \mathbb{R}^{\mathcal{K}\times 2}$ (where ``s'' indicating a single focal agent) by utilizing the k-means clustering algorithm on the endpoints of ground-truth (GT) trajectories in the training dataset (refer to Fig.~\ref{fig:cluster}). 
Each intention point embodies an implicit motion mode, accounting for both motion direction and velocity. 
Given the intention points of a single focal agent, we model each intention query as the learnable positional embedding of a specific intention point:
\begin{align}\label{eq:mtr_intention_query}
E^{\text{(s)}}_{\text{I}}{\scriptstyle [i]} = \text{MLP}\left(\text{PE}(I^{\text{(s)}}{\scriptstyle [i]})\right),
\end{align}
where $i\in \{1, \dots, \mathcal{K}\}$ and $E_{\text{I}}^{\text{(s)}} \in \mathbb{R}^{\mathcal{K}\times D}$. $\text{PE}(\cdot)$ denotes the sinusoidal position encoding.
Notably, each intention query is responsible for predicting trajectories for a specific motion mode, which stabilizes the training process and facilitates multimodal trajectory prediction since each motion mode possesses its own learnable embedding.
Owing to their learnable and adaptive properties, we require only a minimal number of queries (e.g., 64 queries in our setting) for efficient intention localization, rather than employing densely-placed goal candidates~\cite{zhao2020tnt,gu2021densetnt} to cover the agents' destinations.

\myparagraph{Scene Context Aggregation with Intention Query.}
These intention queries are considered as the learnable query embedding of the transformer decoder layer for aggregating context features from the encoded agent features and map features. 
Specifically, in each transformer decoder layer, we first apply the self-attention module to propagate information among $\mathcal{K}$ intention queries  as follows:
\begin{align}\label{eq:mtr:decoder_sa}
\begin{split}
{F'}_{\text{I}}^{(\text{s})}{\scriptstyle [i]}=\text{MHSA}\bigl(&\text{Q:~}F_{\text{I}}^{(\text{s})}{\scriptstyle [i]} + E_{\text{I}}^{\text{(s)}}{\scriptstyle [i]}, \\
&\text{K:~}\bigl\{F_{\text{I}}^{(\text{s})}{\scriptstyle [j]} + E_{\text{I}}^{\text{(s)}}{\scriptstyle [j]}\bigr\}_{j=1}^{\mathcal{K}},\\ 
&\text{V:~}\bigl\{F_{\text{I}}^{(\text{s})}{\scriptstyle [j]}\bigr\}_{j=1}^{\mathcal{K}}\bigr), 
\end{split}
\end{align}
where $i\in \{1, \dots, \mathcal{K}\}$. $F_{\text{I}}^{(\text{s})}\in\mathbb{R}^{\mathcal{K}\times D}$ is the 
query content feature from the previous transformer decoder layer, and it is initialized as zero in the first transformer decoder layer.  ${F'}_{\text{I}}^{(\text{s})}\in\mathbb{R}^{\mathcal{K}\times D}$ indicates the updated query content feature. 
Next, to aggregate scene context features from the encoder network, inspired by \cite{meng2021conditional,liu2022dabdetr},  we concatenate content features and position embedding for both query and key to decouple their contributions to the attention weights. 
Thus, the cross-attention module can be formulated as follows:
\begin{align}\label{eq:mtr:decoder_ca}
\small
\begin{split}
{F''}_{\text{I}}^{(\text{s})}{\scriptstyle [i]}=\text{MHCA}\bigl(&\text{Q:~}[{F'}_{\text{I}}^{(\text{s})}{\scriptstyle [i]}, E_{\text{I}}^{\text{(s)}}{\scriptstyle [i]}], \\
&\text{K:~} [F_{\text{A}}^{\text{(g)}}, \text{PE} (P_\text{A}^{\text{(g)}})] \cup 
[F_{\text{M}}^{\text{(g)}}, \text{PE} (P_\text{M}^{\text{(g)}})],
\\ &\text{V:~}F_{\text{A}}^{\text{(g)}} \cup F_{\text{M}}^{\text{(g)}}\bigr), 
\end{split}
\end{align}
where $i\in \{1, \dots, \mathcal{K}\}$. $\text{MHCA}(\cdot_\text{query}, ~\cdot_\text{key}, ~\cdot_\text{value})$ denotes the multi-head cross-attention layer~\cite{vaswani2017attention}. The sign "$[\cdot, \cdot]$" indicates feature concatenation, and "$\cup$" combines the agent tokens and map tokens as the key and value of the cross-attention module.
Finally, ${F''}_{\text{I}}^{(\text{s})}\in \mathbb{R}^{\mathcal{K}\times D}$ is the final updated query content feature in this transformer decoder layer.

Additionally, for each intention query, we introduce the dynamic map collection strategy to extract fine-grained trajectory features by querying map features from a trajectory-aligned local region.
Specifically, by adopting such a module, the key and value of the map tokens in Eq.~\eqref{eq:mtr:decoder_ca} are restricted to a local region by gathering the polylines whose centers are nearest to the predicted trajectory of the current intention query. 
As the agent's behavior is largely influenced by road maps, this local movement refinement strategy enables a continuous focus on the most recent local context information for iterative motion refinement.

\begin{figure*}[t]
    \centering
    \includegraphics[width=0.99\linewidth]{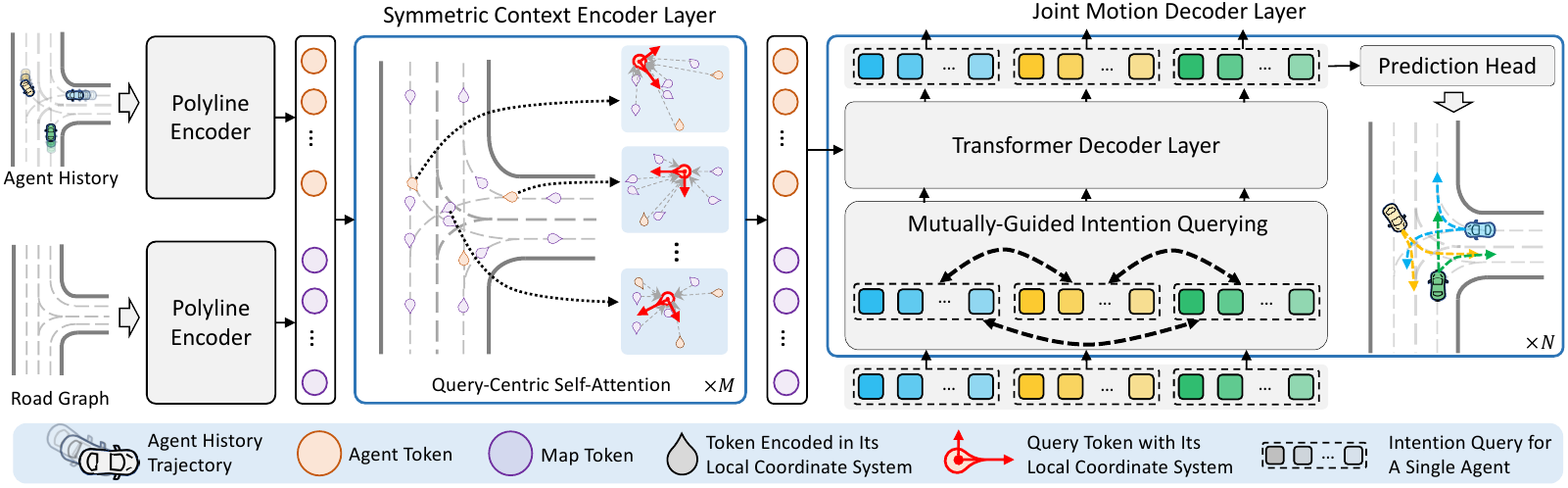}\\
		\vspace{-0.1in}
    \caption{
    The architecture of the MTR++ framework builds upon the initial MTR framework and introduces several enhancements. In the MTR++ framework, we introduce the symmetric context encoder layer, which facilitates the understanding of relationships among tokens within their respective local coordinate systems. By incorporating these symmetrically encoded token features as input, the MTR++ framework employs a joint motion decoder that leverages multiple sets of intention queries. This enables the simultaneous prediction of future trajectories for multiple agents, with the mutually-guided intention querying module facilitating the interaction of future behaviors among different agents. As a result, the MTR++ framework generates more scene-compliant future trajectories, enhancing the overall predictive capabilities of the MTR framework.
}
    \label{fig:mtr++_framework}
\end{figure*}

\myparagraph{Global Intention Localization.}
By considering different motion modes with different learnable queries, the intention queries capture representative features ${F''}_{\text{I}}^{(\text{s})}\in \mathbb{R}^{\mathcal{K}\times D}$ to model the focal agent's future motion. 
Thus, we propose to coarsely localize the agent's intention by predicting the occurrence probability of each intention point as follows: 
\begin{align}\label{eq:gau_reg}
p = \text{MLP}({F''}_{\text{I}}^{(\text{s})}),
\end{align}
where $p\in \mathbb{R}^{\mathcal{K}}$ is a probability distribution to model the potential future intention of the focal agent. 

\myparagraph{Local Movement Refinement.}
To complement the coarse global intention localization, we further predict the detailed future trajectory for each intention query as follows: 
\begin{align}\label{eq:gau_reg}
Z = \text{MLP}({F''}_{\text{I}}^{(\text{s})}),
\end{align}
where $Z\in \mathbb{R}^{\mathcal{K}\times (T \times 5)}$ indicates the $\mathcal{K}$ predicted future trajectories, and each of them has $T$ future frames. 
"5" indicates that we model the uncertainty of each trajectory waypoint with Gaussian distribution as $\mathcal{N}(\mu_x, \sigma_x; \mu_y, \sigma_y; \rho)$. 

As the query content feature ${F''}_{\text{I}}^{(\text{s})}$ will be constantly propagated to the next transformer decoder layer as the new query content feature, the predicted future trajectories can be iteratively refined with multiple stacked transformer decoder layers by continually aggregating scene context features from the encoder network.

\subsection{Multimodal Prediction with Gaussian Mixture Model}\label{sec:mtr_gmm}
As the behaviors of the agents are highly multimodal, we follow \cite{chai2019multipath,varadarajan2021multipath++} to represent the distribution of predicted trajectories with Gaussian Mixture Model (GMM) at each time step.
Specifically, for a specific future time step $i$, MTR will predict $\mathcal{K}$ candidate goal positions with distribution $\mathcal{N}_{1:\mathcal{K}}(\mu_x, \sigma_x; \mu_y, \sigma_y; \rho)$ and probability distribution $p\in \mathbb{R}^{\mathcal{K}}$.
The predicted distribution of the focal agent's position at time step $i$ can be formulated as a GMM with $\mathcal{K}$ components:
\begin{align}\label{eq:gau_prob}
\mathcal{P}_i(o) = \sum_{k=1}^{\mathcal{K}}p_k\cdot f_k(o_x-\mu_x, o_y - \mu_y),
\end{align}
where $f_k(\cdot, \cdot)$ is the probability density function of the $k$-th component of this GMM, and $\mathcal{P}_i(o)$ is the occurrence probability density of the agent at spatial position  $o\in \mathbb{R}^{2}$.
The predicted trajectories can be generated by simply extracting the predicted centers of Gaussian components. 

\myparagraph{Training Loss.}
Given the predicted  Gaussian Mixture Models for a specific future time step, we adopt negative log-likelihood loss to maximize the likelihood of the agent's ground-truth position $(\hat{Y}_x, \hat{Y}_y)$ at this time step, and the detailed loss can be formulated as:
\begin{align}\label{eq:gmm_loss}
L_{\text{GMM}}&=-\log f_h(\hat{Y}_x - \mu_x, \hat{Y}_y - \mu_y) - \text{log}(p_h),
\end{align}
where $f_h(\hat{Y}_x - \mu_x, \hat{Y}_y - \mu_y)$ is the selected positive Gaussian component for optimization. Here the positive Gaussian component is selected by finding the closest intention query with the endpoint of this GT trajectory.
$p_h$ is the predicted probability of this selected positive Gaussian component, and we adopt cross entropy loss in the above equation to maximize the probability of the selected positive Gaussian component.
The final loss of our framework is denoted as: 
\begin{align}
    L_{\text{SUM}} = L_{\text{GMM}} + L_{\text{DMP}}, 
\end{align}
where $L_{\text{DMP}}$ is the $L1$ regression loss on the outputs of Eq.~\eqref{eq:aux_reg}.

\section{MTR++: Multi-Agent Motion Prediction}
The above MTR framework proposed for multimodal motion prediction has demonstrated state-of-the-art performance. However, its scene context modeling module adopts the focal-agent-centric strategy commonly found in previous works~\cite{zhao2020tnt,gu2021densetnt,varadarajan2021multipath++}, which encodes the scene context separately for each focal agent, leading to computational inefficiencies when predicting motion for multiple agents. Although the Scene Transformer model~\cite{ngiam2021scene} presents a shared context encoding strategy for predicting trajectories of multiple agents, it still centers the scene around a specific agent, potentially limiting its performance for off-center agents due to uneven distribution of shared context information.

To address the aforementioned challenges, we introduce an enhanced version of the MTR framework, denoted as MTR++. As shown in Fig.~\ref{fig:mtr++_framework}, the MTR++ framework enables simultaneous motion prediction of multiple agents via shared symmetric scene context modeling and mutually-guided intention querying. We elaborate these two improvements in Sec.~\ref{sec:mtr++_encoder} and Sec.~\ref{sec:mtr++_decoder}, respectively.

\begin{figure}[h]
	\centering
	\includegraphics[width=0.99\linewidth]{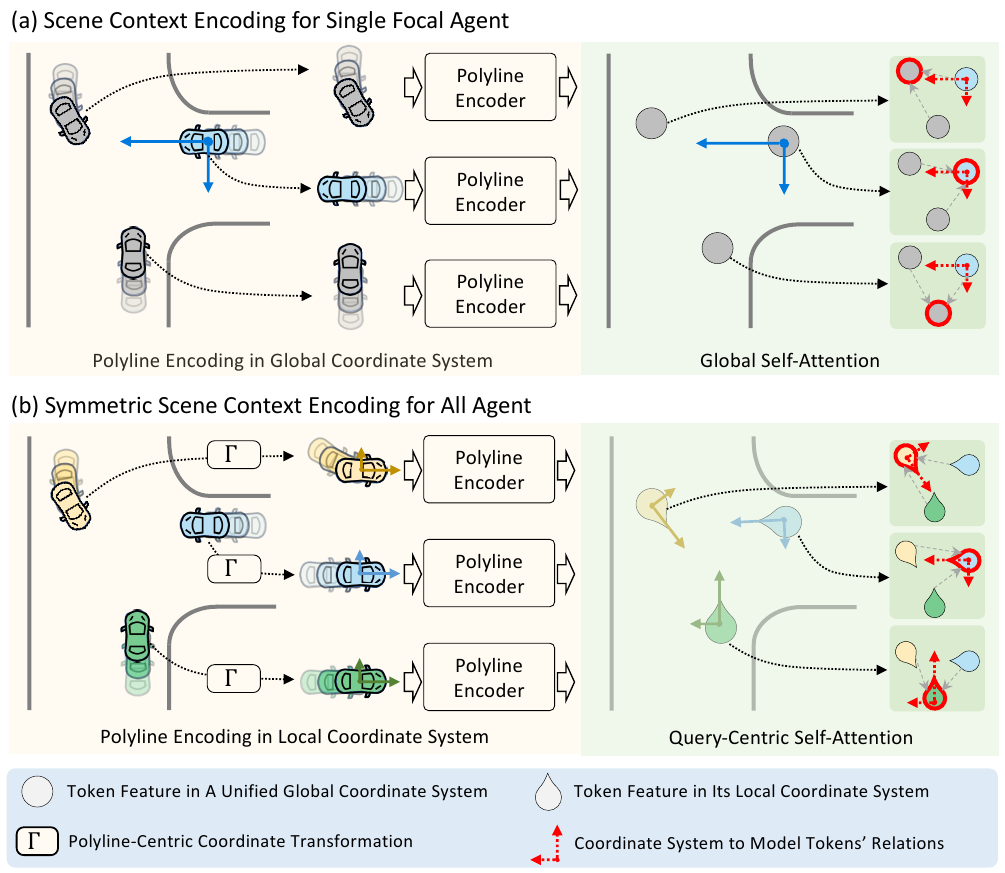}\\
	\vspace{-2mm}
	\caption{The comparison of two different scene context encoding modules in the MTR and MTR++ frameworks. The MTR framework adopts the scene context encoding for a single focal agent, where both the polyline-wise features and tokens' relationship are encoded in a global coordinate system. In contrast, the MTR++ framework encodes both the polyline-wise features and their relationship in their respective local coordinate system via the novel query-centric self-attention module, thus enabling simultaneous motion prediction of multiple agents.}
	\label{fig:context_encoder}
	\vspace{-3mm}
\end{figure}

\subsection{Symmetric Scene Context Modeling for All Agents}\label{sec:mtr++_encoder}

To improve the efficiency of predicting future trajectories of multiple agents simultaneously, we propose a symmetric scene context modeling module that employs a shared context encoder to encode complex multimodal scene context information for all agents. In contrast to most existing methods that center the scene around a particular agent, our approach encodes the entire scene symmetrically for each agent. As a result, the encoded scene context features can be directly utilized for predicting the motion of any agent by attaching a motion decoder network.

\myparagraph{Input Representation with Polyline-Centric Encoding.}
We employ the same vectorized representation as in Sec.~\ref{sec:mtr_encoder} to encode the input context features. However, instead of normalizing all inputs to the global coordinate system centered on one focal agent, we encode the feature of each polyline in a polyline-centric local coordinate system (see Fig.~\ref{fig:context_encoder}). 
Specifically, we modify the polyline feature encoding process in Eq.~\eqref{eq:mtr_polyline}  by incorporating the coordinate transformation function, denoted as $\Gamma(\cdot)$, as follows: 
\begin{align}\label{eq:mtr++_polyline}
\small
\begin{split}
F_{\text{A}}^{(l)} = \phi\left(\text{MLP}\bigl(\Gamma(S_{\text{A}}^{\text{(g)}})\bigr)\right),~~~~
F_{\text{M}}^{(l)} = \phi\left(\text{MLP}\bigl(\Gamma(S_{\text{M}}^{\text{(g)}})\bigr)\right),
\end{split}
\end{align}
where $\Gamma(\cdot)$ transforms the polyline features from an arbitrary global coordinate system to the polyline-centric local coordinate system. Concretely, we use the latest position and moving direction of each agent to determine the local coordinate system of their corresponding polyline, while for the map polylines, we calculate the geometry center and tangent direction of each polyline to determine their local coordinate system. 

The encoded features $F_{\text{A}}^{(l)}\in \mathbb{R}^{N_a\times D}$ and  $F_{\text{M}}^{(l)} \in \mathbb{R}^{N_m\times D}$ (where ``$l$'' indicating the local reference frame) capture the polyline-wise features for the agent history states and map polylines, respectively. 
Importantly, these polyline features are encoded in their own local coordinate system, independent of any global coordinate system.  This provides input token features that are decoupled from the global coordinate system and enables the symmetric modeling of token relations in the subsequent step.

\myparagraph{Attribute Definition of Polyline Tokens.}
The features $F_{\text{A}}^{(l)}$ and $F_{\text{M}}^{(l)}$ are considered as input tokens in the subsequent transformer network, and their features are concatenated to form the input token feature matrix ${F_{\text{AM}}^{(l)}}=[F_{\text{A}}^{(l)}, F_{\text{M}}^{(l)}] \in \mathbb{R}^{(N_a+N_m)\times D}$. As in Sec.~\ref{sec:mtr_encoder}, the global positions of these tokens are denoted as $P_{\text{AM}}^{\text{(g)}} \in \mathbb{R}^{(N_a+N_m)\times 2}$, which can be defined in an arbitrary global coordinate system. Additionally, each token is associated with a heading direction attribute $H_{\text{AM}}^{\text{(g)}} \in \mathbb{R}^{(N_a+N_m)\times 1}$, which is defined similarly to the direction definition as in the transformation function $\Gamma(\cdot)$ presented in Eq.~\eqref{eq:mtr++_polyline}.

\myparagraph{Symmetric Scene Context Modeling with Query-Centric Self-Attention.}
In our previous MTR framework, we model the relationship between the input token features using a self-attention module (Eq.~\eqref{eq:mtr_encoder_sa}) that depends on a global coordinate system centered on a single focal agent. However, this approach hindered the performance of motion prediction for other agents. To address this limitation, we propose a \textit{query-centric self-attention} module, which models the relationship between all tokens in a symmetric manner, decoupled from any global coordinate system.

Specifically, to explore the relationship between a query token and other tokens in its specific local coordinate system, we perform the attention mechanism separately for each query token. For instance, let us consider the $i$-th token as the query. We convert the coordinates and directions of all tokens into the local coordinate system of the query token:
\begin{align}\label{eq:mtr++_encoder_rel_pos}
\small
R_{\text{AM}}^{(\text{pos})}{\scriptstyle [i,j]} &= (P_{\text{AM}}^{(\text{g})}{\scriptstyle [j]}-P_{\text{AM}}^{(\text{g})}{\scriptstyle [i]}) \begin{bmatrix}
    \cos{H_{\text{AM}}^{(\text{g})}{\scriptstyle [i]}} & -\sin{H_{\text{AM}}^{(\text{g})}{\scriptstyle [i]}} \\
    \sin{H_{\text{AM}}^{(\text{g})}{\scriptstyle [i]}} & \cos{H_{\text{AM}}^{(\text{g})}{\scriptstyle [i]}} \\
\end{bmatrix},\nonumber\\ 
R_{\text{AM}}^{(\text{ang})}{\scriptstyle [i,j]} &= H_{\text{AM}}^{(\text{g})}{\scriptstyle [j]}-H_{\text{AM}}^{(\text{g})}{\scriptstyle [i]},  
\end{align}
where $i\in \{1, \dots, N_a+N_m\}$, and $j\in \Omega(i)$  indicating the index of its neighboring tokens.
$R_{\text{AM}}^{(\text{pos})}{\scriptstyle [i,j]} \in \mathbb{R}^2$ and $R_{\text{AM}}^{(\text{ang})}{\scriptstyle [i,j]} \in \mathbb{R}$ indicate the $j$-th token's relative position and direction in the local coordinate system of the $i$-th query token. 
We then perform the query-centric self-attention mechanism as follows:
\begin{align}\label{eq:mtr++_encoder_sa}
\begin{split}
{F'}^{(l)}_{\text{AM}}{\scriptstyle [i]}=\text{MHSA}\bigl(&\text{Q:~}[F_{\text{AM}}^{(l)}{\scriptstyle [i]},  \text{PE}(R_{\text{AM}}{\scriptstyle [i,i]})], \\
&\text{K:~}\bigl\{[F_{\text{AM}}^{(l)}{\scriptstyle [j]}, \text{PE}(R_{\text{AM}}{\scriptstyle [i,j]})]\bigr\}_{j\in \Omega(i)},\\ 
&\text{V:~}\bigl\{F_{\text{AM}}^{(l)}{\scriptstyle [j]} + \text{PE}(R_{\text{AM}}{\scriptstyle [i,j]})\bigr\}_{j\in \Omega(i)}\bigr), 
\end{split}
\end{align}
where PE$(R_{\text{AM}}{\scriptstyle [i,j]})$ indicates the sinusoidal positional encoding of both $R^{(\text{pos})}_{\text{AM}}{\scriptstyle [i,j]}$ and $R^{(\text{ang})}_{\text{AM}}{\scriptstyle [i,j]}$. 
This query-centric self-attention mechanism models the token relationship in a symmetric manner by integrating the  global-coordinate-decoupled token feature $F_{\text{AM}}^{(l)}$ and relative coordinate $R_{\text{AM}}$ based on the query token.

Note that the computational cost of the proposed query-centric self-attention module in Eq.\eqref{eq:mtr++_encoder_sa} is comparable to that of the self-attention module in Eq.\eqref{eq:mtr_encoder_sa}. The key advantage of the proposed module is that it enables the symmetric encoding of scene context features for each input token, such as each agent, thus allowing the encoded features to be used for predicting the motion of any input agent. This feature enables a shared scene context encoder for simultaneous prediction of the motion of multiple agents.

\begin{figure}[h]
	\centering
	\includegraphics[width=0.99\linewidth]{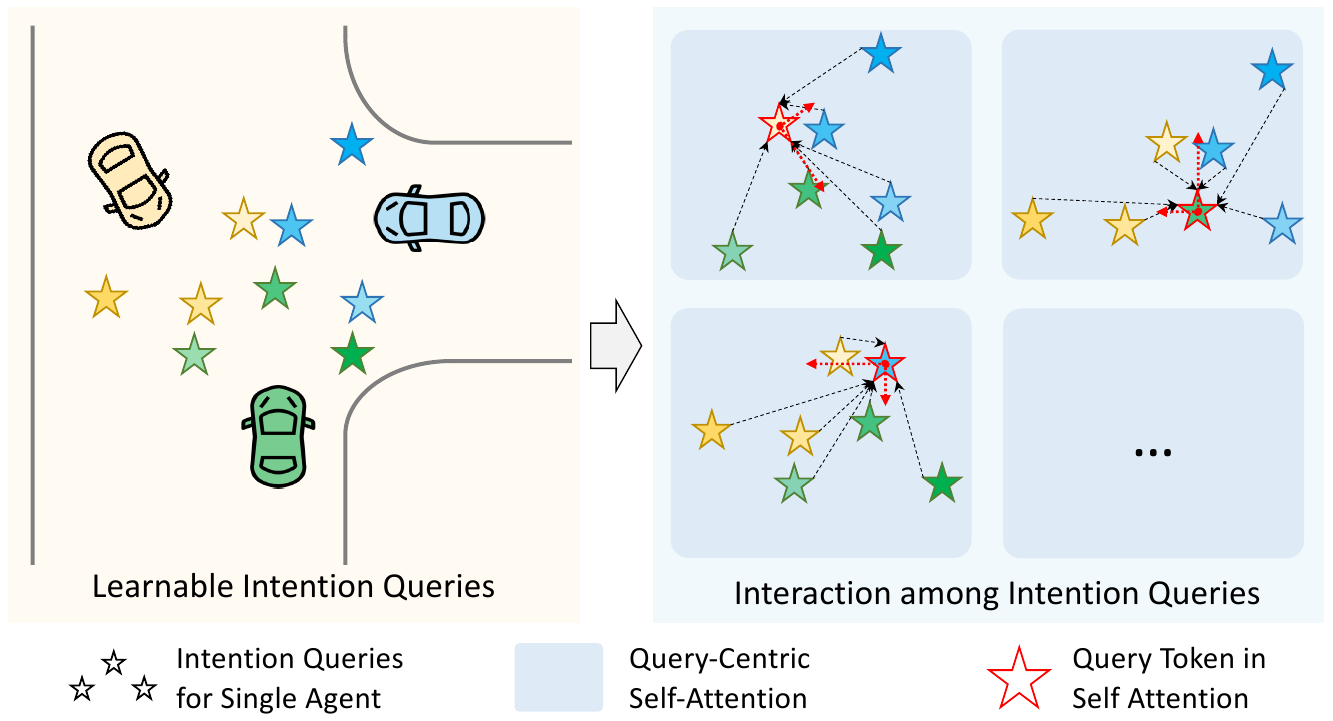}\\
	\vspace{-2mm}
	\caption{The illustration of the mutually-guided intention querying module.}
	\label{fig:mutually_guided_intention_query}
	\vspace{-3mm}
\end{figure}

\subsection{Joint Motion Decoder with Mutually-Guided Queries}\label{sec:mtr++_decoder}
The MTR++ framework utilizes symmetrically encoded scene context features, which are fed to the motion decoder as described in Sec.~\ref{sec:mtr_decoder}, to enable simultaneous motion prediction for multiple focal agents. This simultaneous motion prediction allows for the exploration of future behavior interactions among the agents, which is crucial for making more accurate and scene-compliant motion predictions. 

\myparagraph{Mutually-Guided Intention Querying of Multiple Agents.}
To enhance the accuracy of motion prediction by enabling agents to interact and influence each other's behavior, as shown in Fig.~\ref{fig:mutually_guided_intention_query}, we propose a \textit{mutually-guided intention querying} module. 
However, building such interaction is non-trivial since the intention queries of different focal agents are encoded in their own local coordinate system as in Eq.~\eqref{eq:mtr_intention_query}.  
To maintain the local-encoded features of intention queries while also establishing the spatial relationship among them, we adopt the previously introduced query-centric self-attention module, similar to the one used in Sec.~\ref{sec:mtr++_encoder}, to enable the information interaction among all intention queries.

Specifically,  to predict future trajectories for $N_o$ focal agents, the motion decoding process is conducted simultaneously in the local coordinate system centered at each focal agent. The intention queries for the focal agents are represented as $E_{\text{I}}^{(\text{m})} \in \mathbb{R}^{N_o \times \mathcal{K} \times D}$ (where ``m'' indicating multiple focal agents), wherein the intention queries for different focal agents are encoded using Eq.~\eqref{eq:mtr_intention_query} with the same intention points $I^{(\text{s})} \in \mathbb{R}^{\mathcal{K} \times 2}$.

However, as the intention points are defined in their respective local coordinate systems, in order to facilitate information propagation among the intention queries of different focal agents, we first transform their intention points into the same global coordinate system based on the global positions $P_{\text{O}}^{\text{(g)}} \in \mathbb{R}^{N_o \times 2}$ and moving directions $H_{\text{O}}^{\text{(g)}} \in \mathbb{R}^{N_o \times 1}$ of the focal agents, as follows:
\begin{align}
\begin{split}
P_{\text{I}}^{(\text{m})}{\scriptstyle [t]} &= I^{(\text{s})} \begin{bmatrix}
    \cos{H_{\text{O}}^{\text{(g)}}{\scriptstyle [t]}} & \sin{H_{\text{O}}^{\text{(g)}}{\scriptstyle [t]}} \\
    -\sin{H_{\text{O}}^{\text{(g)}}{\scriptstyle [t]}} & \cos{H_{\text{O}}^{\text{(g)}}{\scriptstyle [t]}} \\
\end{bmatrix} + P_{\text{O}}^{\text{(g)}}{\scriptstyle [t]},\\ 
\end{split}
\end{align}
where $t\in \{1, \dots, N_o\}$ and $P_{\text{I}}^{(\text{m})} \in \mathbb{R}^{N_o\times \mathcal{K}\times 2}$. 
To build the information interaction among all intention queries of multiple agents, we re-organize the intention points and intention queries as $P_{\text{I}}^{(\text{m})} \in \mathbb{R}^{(N_o\mathcal{K})\times 2}$ and $E_{\text{I}}^{(\text{m})} \in \mathbb{R}^{(N_o\mathcal{K}) \times D }$, respectively. Meanwhile, we also assign the heading direction $H_{\text{I}}^{(\text{m})} \in \mathbb{R}^{(N_o\mathcal{K})\times 1}$ for the intention queries for calculating their relative spatial relationship, where the $\mathcal{K}$ intention queries of the $t$-th focal agent share the same heading direction as its moving direction $H_{\text{O}}^{\text{(g)}}{\scriptstyle [t]}$. 

Thus, following Eq.~\eqref{eq:mtr++_encoder_rel_pos}, when considering the $i$-th intention query as the query token, we transform the coordinates and directions of all intention queries to the local coordinate system of the $i$-th query token, as follows:
\begin{align}
\small
R_{\text{I}}^{(\text{pos})}{\scriptstyle [i,j]} &= (P_{\text{I}}^{(\text{m})}{\scriptstyle [j]}-P_{\text{I}}^{(\text{m})}{\scriptstyle [i]}) \begin{bmatrix}
    \cos{H_{\text{I}}^{(\text{m})}{\scriptstyle [i]}} & -\sin{H_{\text{I}}^{(\text{m})}{\scriptstyle [i]}} \\
    \sin{H_{\text{I}}^{(\text{m})}{\scriptstyle [i]}} & \cos{H_{\text{I}}^{(\text{m})}{\scriptstyle [i]}} \\
\end{bmatrix},\nonumber\\ 
R_{\text{I}}^{(\text{ang})}{\scriptstyle [i,j]} &= H_{\text{I}}^{(\text{m})}{\scriptstyle [j]}-H_{\text{I}}^{(\text{m})}{\scriptstyle [i]},  
\end{align}
where $i\in \{1, \dots, N_o\mathcal{K}\}$, and $j\in \Omega(i)$ indicating the index of its neighboring tokens.
Then, we apply the query-centric self-attention module on all intention queries as follows: 
\begin{align}\label{eq:mtr++_decoder_sa}
\small
{F'}_{\text{I}}^{(\text{m})}{\scriptstyle [i]}=\text{MHSA}\bigl(&\text{Q:~}[F_{\text{I}}^{(\text{m})}{\scriptstyle [i]} + E_{\text{I}}^{(\text{m})}{\scriptstyle [i]},  \text{PE}(R_{\text{I}}{\scriptstyle [i,i]})], \\
&\text{K:~}\{[F_{\text{I}}^{(\text{m})}{\scriptstyle [j]} + E_{\text{I}}^{(\text{m})}{\scriptstyle [j]}, \text{PE}(R_{\text{I}}{\scriptstyle [i,j]})]\}_{j\in \Omega(i)},\nonumber\\ 
&\text{V:~}\{F_{\text{I}}^{(\text{m})}{\scriptstyle [j]} + E_{\text{I}}^{(\text{m})}{\scriptstyle [j]} + \text{PE}(R_{\text{I}}{\scriptstyle [i,j]})\}_{j\in \Omega(i)}\bigr), \nonumber
\end{align}
where $i\in \{1, \dots, N_o\mathcal{K}\}$, and $F_{\text{I}}^{(\text{m})} \in \mathbb{R}^{(N_o\mathcal{K}) \times D}$ indicates the query content feature from the previous transformer decoder layer and is initialized as zero in the first decoder layer.

Finally, the updated query content feature ${F'}_{\text{I}}^{(\text{m})} \in \mathbb{R}^{N_o \times \mathcal{K}\times D}$ will be utilized individually for the subsequent scene context aggregation of each focal agent. This aggregation process is the same as described in Eq.~\eqref{eq:mtr:decoder_sa} and Eq.~\eqref{eq:mtr:decoder_ca} in the MTR framework. It is worth noting that the positional encoding for the encoded scene elements from the context encoder is defined in the local coordinate system of each focal agent. These resulting query features are then fed into the prediction head, which generates future trajectories for each focal agent.
By establishing this information propagation process, the intention queries of multiple agents are guided by each other during the multimodal motion decoding process, ultimately resulting in more informed and realistic predictions of their future trajectories.

\begin{table*}
\centering\small
    \begin{tabular}{c|l|c||ccccc}
        \hline
        \rowcolor{mygray} & Method & Reference & minADE~$\downarrow$ & minFDE~$\downarrow$ & Miss Rate~$\downarrow$ & {\bf mAP}~$\uparrow$ \\
        \hline
         \multirow{7}{*}{Test} & MotionCNN~\cite{konev2021motioncnn} & CVPRw 2021 & 0.7400 & 1.4936 & 0.2091 & 0.2136  \\
        & ReCoAt~\cite{huangrecoat} & CVPRw 2021 & 0.7703 & 1.6668 & 0.2437 & 0.2711	 \\
        & DenseTNT~\cite{gu2021densetnt} & ICCV 2021 & 1.0387 & 1.5514 & 0.1573 & 0.3281  \\
        & SceneTransformer~\cite{ngiam2021scene}~~~~~~~ & ICLR 2022 & 0.6117 & { 1.2116} & 0.1564 & 0.2788 \\ 
        & HDGT~\cite{jia2022hdgt} & Arxiv 2022 & 0.5933 & 1.2055 & 0.1511 & 0.2854 \\ 
        & MTR (Ours) & NeurIPS 2022 & {0.6050} & 1.2207 & {0.1351} &{0.4129}	 \\
        & MTR++ (Ours) & - & \textbf{0.5906}	& \textbf{1.1939} &	\textbf{0.1298} & \textbf{0.4329}\\
        \cline{2-7} 
        & $^\dagger$MultiPath++~\cite{varadarajan2021multipath++} & ICRA 2022 & \textit{0.5557} & \emph{1.1577}
        & \emph{0.1340} & \emph{0.4092}  \\
        & $^\dagger$MTR++\_Ens (Ours) & - & \emph{0.5581} & \textit{1.1166} & \textit{0.1122} & \textit{0.4634}	 \\
        \hline 
        \multirow{2}{*}{Val} & MTR (Ours) & NeurIPS 2022 & 0.6046 & 1.2251 & 0.1366 & 0.4164 \\
        & MTR++ (Ours) & - & \textbf{0.5912} & \textbf{1.1986} & \textbf{0.1296} & \textbf{0.4351} \\
        \hline
\end{tabular}
\vspace{1mm}
\caption{Performance comparison of marginal motion prediction on the test and validation set of Waymo Open Motion Dataset. 
$\dagger$: The results are shown in \emph{italic} for reference since their performance is achieved with model ensemble techniques. }
\vspace{-5mm}
\label{tab:wod_test}
\end{table*}

\begin{table*}
    \centering
    \small
        \begin{tabular}{c|l|c||ccccc}
            \hline
            \rowcolor{mygray} & Method & Reference & minADE~$\downarrow$ & minFDE~$\downarrow$ & Miss Rate~$\downarrow$ & { \bf mAP}~$\uparrow$ \\
            \hline
            \multirow{7}{*}{Test}& Waymo LSTM baseline~\cite{ettinger2021large} & ICCV 2021 & 1.9056 & 5.0278 & 0.7750 & 0.0524  \\
            & HeatIRm4~\cite{mo2021multi} & CVPRw 2021 & 1.4197 & 3.2595 & 0.7224 & 0.0844 \\
            & AIR$^2$~\cite{wu2021air} & CVPRw 2021 & 1.3165 & 2.7138 & 0.6230 & 0.0963 \\ 
            & SceneTransformer~\cite{ngiam2021scene} & ICLR 2022 & 0.9774 & 2.1892 & 0.4942 & 0.1192 \\ 
            & M2I~\cite{sun2022m2i} & CVPR 2022 & 1.3506 & 2.8325 & 0.5538 & 0.1239\\ 
            & MTR (Ours) & NeurIPS 2022 & {0.9181} & { 2.0633} & { 0.4411}	 & { 0.2037}	\\
            & MTR++ (Ours) & - & \textbf{0.8795} & \textbf{1.9509} & \textbf{0.4143} & \textbf{0.2326}	\\ 
            \hline 
            \multirow{2}{*}{Val}& MTR (Ours) & NeurIPS 2022 & 0.9132 & 2.0536	& 0.4372 & 0.1992\\
            & MTR++ (Ours) & - & \textbf{0.8859} &	\textbf{1.9712} & \textbf{0.4106} & \textbf{0.2398}	 \\ 
            	
            \hline
    \end{tabular}
    \vspace{1mm}
    \caption{Performance comparison of joint motion prediction on the interactive validation and test set  of Waymo Open Motion Dataset.}
    \vspace{-5mm}
    \label{tab:wod_interactive}
\end{table*}
    
\section{Experiments}

\subsection{Experimental Setup}\label{sec:exp_setup}
\myparagraph{Dataset and metrics.}
We mainly evaluate our approach using the Waymo Open Motion Dataset (WOMD)\cite{ettinger2021large}, a large-scale dataset that captures diverse traffic scenes with interesting interactions among agents. 
There are two tasks in WOMD with separate evaluation metrics: 
(1) The \emph{marginal motion prediction challenge} that independently evaluates the predicted motion of each agent (up to 8 agents per scene). 
(2) The \emph{joint motion prediction challenge} that needs to predict the joint future positions of 2 interacting agents for evaluation. 
 For both tasks, the dataset provides 1 second of history data and aims to predict 6 marginal or joint trajectories of the agents for 8 seconds into the future.
The dataset contains $487k$ training scenes, and approximately $44k$ validation scenes and $44k$ testing scenes for each challenge. We utilize the official evaluation tool, which calculates important metrics such as mAP and miss rate, as used in the official WOMD leaderboards~\cite{womd_leaderboard2023,womd_leaderboard_interact}.

In addition to the WOMD, we also evaluate our approach on the Argoverse 2 Motion Forecasting Dataset~\cite{wilson2021argoverse}, another large-scale motion prediction dataset. It contains 250,000 scenarios for training and validation. The model needs to take the history five seconds of each scenario as input and predict the six-second future trajectories of one interested agent, where HDMap is always available to provide map context information. We also utilize the official evaluation tool to calculate the miss rate as the main metric. 

\myparagraph{Implementation details.}
For both the MTR and MTR++ frameworks, we stack 6 transformer encoder layers for context encoding. 
The road map is represented as multiple polylines, where each polyline contains up to 20 map points (about $10m$ in WOMD). 
We select $N_m=768$ nearest map polylines around the interested agents. The number of neighbors in the encoder's local self-attention is set to 16. 
The hidden feature dimension is set as $D=256$.
For the decoder modules, we stack 6 decoder layers. 
For dynamic map collection, we collect the closest 128 map polylines from the context encoder for iterative motion refinement. 
By default, we utilize 64 motion query pairs where their intention points
are generated by conducting the k-means clustering algorithm on the training dataset. 
The number of neighbors for the query-centric self-attention module is set to 16 for the MTR++ framework.
To generate 6 future trajectories for evaluation, we use non-maximum suppression (NMS) to select the top 6 predictions from 64 predicted trajectories by calculating the distances between their endpoints, and the distance threshold is set as $2.5m$. More implementation details of the initial MTR framework can be found in our open-source codebase: \url{https://github.com/sshaoshuai/MTR}.

\myparagraph{Training details.}
Our model is trained in an end-to-end manner by AdamW optimizer with a learning rate of 0.0001, a weight decay of 0.01, and a batch size of 80 scenes. 
We train the model for 30 epochs with 8 GPUs, and the learning rate is decayed by a factor of 0.5 every 2 epochs from epoch 20.

\subsection{Main Results}
\myparagraph{Performance comparison for marginal motion prediction.}
We evaluate the marginal motion prediction performance of our MTR frameworks by comparing them with leading-edge research on the WOMD test set. As presented in Table~\ref{tab:wod_test}, our initial MTR framework already surpasses previous state-of-the-art approaches~\cite{gu2021densetnt,ngiam2021scene,jia2022hdgt} with significant improvements. It achieves an mAP increase of $+8.48\%$ and reduces the miss rate from $15.11\%$ to $13.51\%$.
Furthermore, our latest MTR++ framework further enhances the performance compared to MTR on all metrics. Particularly, it achieves a $+2.00\%$ improvement in mAP, showcasing its ability to generate more confident multimodal future trajectories by jointly considering the future behaviors of multiple agents. 

Additionally, we also adopt a simple model ensemble strategy, combining predictions from multiple models and employing non-maximum-suppression (NMS) to remove redundant predictions. By adopting this ensemble strategy to diverse variants of our MTR frameworks (\emph{e.g.}, more decoder layers, different number of queries, larger hidden dimension), our approach significantly outperforms the previous state-of-the-art ensemble result~\cite{varadarajan2021multipath++}, increasing the mAP by $+5.42\%$ and reducing the miss rate from $13.40\%$ to $11.22\%$. 

Notably, our MTR and MTR++ frameworks have secured the first-place positions in the highly-competitive Waymo Motion Prediction Challenge in 2022~\cite{womd_leaderboard2022} and 2023~\cite{womd_leaderboard2023}, respectively. As of May 30, 2023, our MTR++ framework holds the $1^{st}$ rank on the motion prediction leaderboard of WOMD~\cite{womd_leaderboard2023}, outperforming other works by a significant margin. These notable achievements highlight the effectiveness of the MTR frameworks.

\begin{table}
    \centering\small
    \resizebox{0.49\textwidth}{!}{
        \begin{tabular}{l|cc|cc}
            \hline
            \rowcolor{mygray}\tabincell{c}{Method}  & \tabincell{c}{Number of \\Parameters} & \tabincell{c}{Inference\\ Latency} &  Miss Rate~$\downarrow$ & {\bf mAP}~$\uparrow$ \\
            \hline 
            $^\dagger$SceneTransformer~\cite{ngiam2021scene} & 15.3M & 52ms (V100) & 0.1564 &  0.2788	\\
            DenseTNT~\cite{gu2021densetnt} & 1.1M & 540ms & 0.1573 &  0.3281\\
            HDGT~\cite{jia2022hdgt} & 12.1M & 1320ms & 0.1511 & 0.2854	 \\
            \hline 
            MTR++ (light) & 11.7M & 67ms & 0.1430 & 0.3896 \\
            MTR & 65.8M & 193ms & 0.1351 & 0.4129  \\  
            MTR++ & 86.6M & 118ms & \textbf{0.1298} & \textbf{0.4329} \\  
            \hline
        \end{tabular}
    }
    \vspace{1mm}
    \caption{Performance is compared in terms of the number of parameters and inference latency for multi-agent prediction, utilizing the Waymo Open Motion Dataset. Latency is estimated by mandating each model to predict the motion of 32 agents per scenario, leveraging an NVIDIA Quadro RTX 8000 GPU. $\dagger$: the number of parameters and latency are reported in the respective source paper, using NVIDIA V100 GPU.}
    \label{tab:latency_params}
    \vspace{-5mm}
\end{table}

\begin{figure*}
		\centering
		\includegraphics[width=0.99\linewidth]{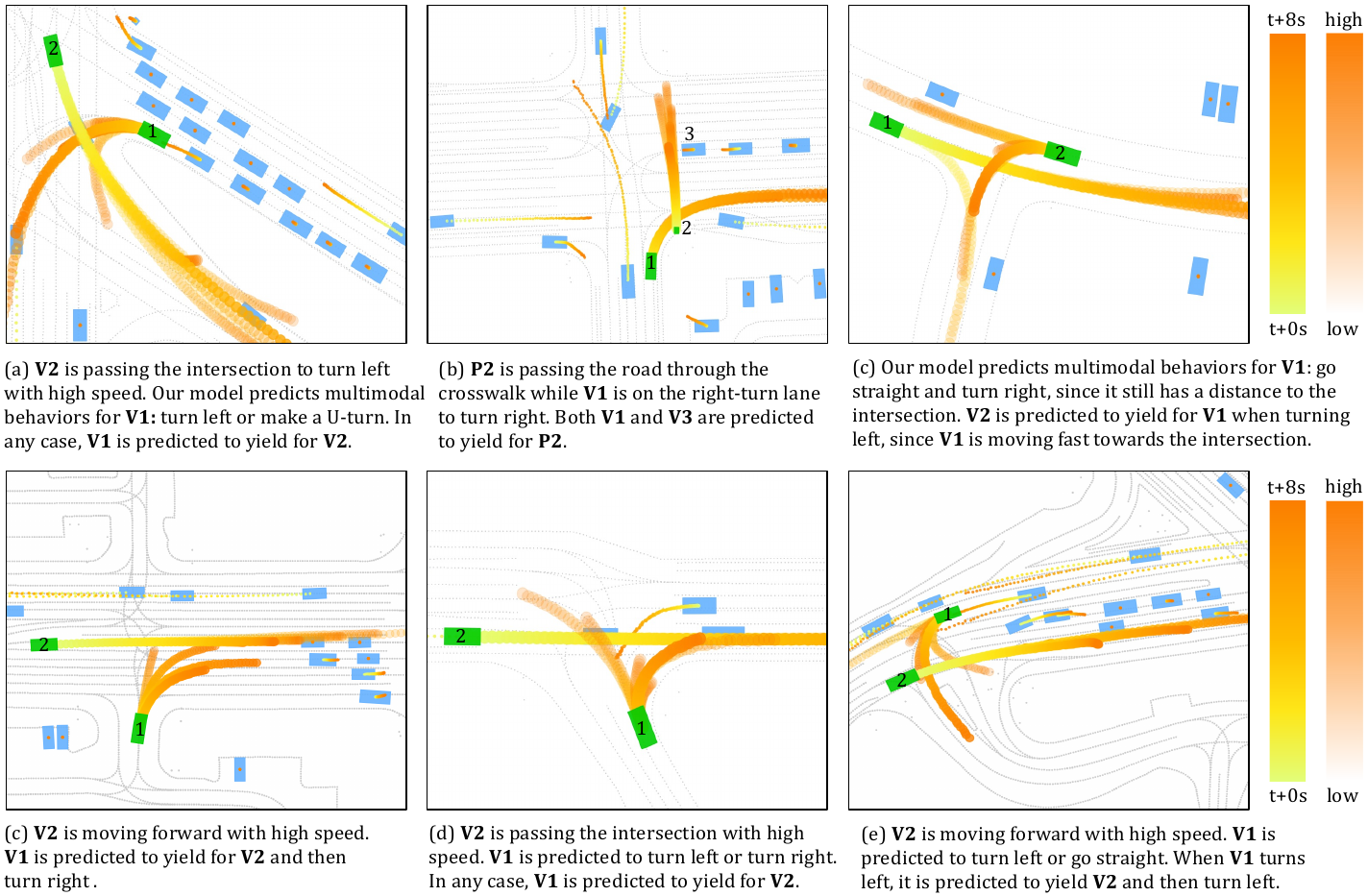}
		\vspace{-2mm}
		\caption{Qualitative results of MTR frameworks on WOMD. There are two interested agents in each scene (green rectangle), where our model predicts 6 multimodal future trajectories for each of them. For other agents (blue rectangle), a single trajectory is predicted by the dense future prediction module. We use gradient color to visualize the trajectory waypoints at different future time steps, and trajectory confidence is visualized by setting different transparent. Abbreviation: Vehicle (V), Pedestrian (P).}
		\label{fig:demo}
\end{figure*}

\myparagraph{Performance comparison for joint motion prediction.}
We also evaluate the proposed MTR frameworks on the joint motion prediction benchmark, merging the marginal predictions of two interacting agents into a joint prediction as explained in \cite{casas2020implicit,ettinger2021large,sun2022m2i}. We select the top 6 joint predictions from 36 potential combinations of these agents, with the confidence of each combination being the product of marginal probabilities.
As indicated in Table~\ref{tab:wod_interactive}, our initial MTR framework already surpasses state-of-the-art approaches~\cite{ngiam2021scene,sun2022m2i} by substantial margins on all measures, reducing the miss rate from $49.42\%$ to $44.11\%$ and enhancing the mAP from $12.39\%$ to $20.37\%$.
Furthermore, our advanced MTR++ framework, which allows us to concurrently predict future motion for two interactive agents with shared context encoding, amplifies the robust performance of MTR across all metrics, achieving an increase of $+2.89\%$ in terms of mAP and reducing the miss rate by $2.68\%$. The extraordinary performance enhancements of the MTR++ framework emphasize that, through the adoption of symmetric scene context encoding and mutually-guided intention querying, our MTR++ framework can accurately predict future trajectories that exhibit scene consistency among highly interacting agents. Additionally, we also provide some qualitative results in Fig.~\ref{fig:demo} to show our predictions in complicated interacting scenarios.
Notably, as of May 30, 2023, our MTR++ framework holds the  $1^{st}$ rank on the joint motion prediction leaderboard of WOMD~\cite{womd_leaderboard_interact}. 

Moreover, we provide the performance comparison in terms of the number of parameters and inference latency in Table~\ref{tab:latency_params}. 
To enable a comparison of performance and latency with a similar parameter count, we have provided a streamlined version of MTR++ (notated as ``MTR++ (light)") by diminishing the number of encoder layers to 2 and decoder layers to 3.
As indicated in Table~\ref{tab:latency_params}, our lightweight MTR++ surpasses existing approaches markedly, enhancing the mAP from $32.81\%$ to $38.96\%$. When compared with SceneTransformer~\cite{ngiam2021scene}, our MTR++ (light) not only employs fewer parameters ($15.3$M down to $11.7$M) but also attains superior performance ($27.88\%$ boosting to $38.96\%$). When compared with DenseTNT~\cite{gu2021densetnt}, albeit our MTR++ (light) utilizes more parameters, it operates considerably faster (67ms versus 540ms). This enhanced speed can be attributed to our method sidestepping the densely sampled goal points present in DenseTNT. When compared with HDGT~\cite{jia2022hdgt}, our MTR++ (light) runs significantly faster (67ms as opposed to 1320ms). This efficiency arises from our usage of the well-optimized native Transformer module, whereas HDGT deploys a custom graph neural network. Furthermore, our standard MTR and MTR++ elevate performance through the use of additional parameters, while simultaneously maintaining inference latencies significantly lower than the existing open-source approaches~\cite{gu2021densetnt, jia2022hdgt}.

\myparagraph{Performance comparison on the Argoverse 2 dataset.}
As shown in Table~\ref{tab:argo2_test}, we also provide the performance comparison of our approach on the Argoverse 2 dataset for reference. 
We compare our approach with the top-10 submissions on the leaderboard of Argoverse 2 dataset~\cite{argo2_leaderboard} at the time of our MTR framework submission. These submissions, primarily developed for the Argoverse 2 Motion Forecasting Competition 2022, represent highly competitive approaches. Notably, our MTR framework achieves new state-of-the-art performance with remarkable improvements in miss-rate-related metrics. 
Moreover, our MTR++ further surpasses the performance of both MTR and other existing approaches across all three metrics, thereby highlighting the exceptional generalizability and robustness of our approach.

\begin{table}[t]
    \centering\small
    \setlength\tabcolsep{10pt}
	\resizebox{0.48\textwidth}{!}{
        \begin{tabular}{c|ccc}
            \hline
            \rowcolor{mygray}Method & \tabincell{c}{Miss Rate $\downarrow$\\($K$=6)} & \tabincell{c}{Miss Rate $\downarrow$\\($K$=1)} & \tabincell{c}{brier-minFDE $\downarrow$ \\ ($K$=6)} \\ 
            \hline
            MTR++ (Ours) & {\bf 0.14} & {\bf 0.56} & {\bf 1.88} \\
            MTR (Ours) & 0.15 & 0.58 & 1.98 \\
        TENET~\cite{wang2022tenet} & 0.19 & 0.61 & 1.90\\
            OPPred & 0.19 & 0.60 & 1.92\\
            Qml & 0.19 & 0.62 & 1.95\\
            GANet & 0.17 & 0.60 & 1.97\\
            VI LaneIter & 0.19 & 0.61 & 2.00\\
            QCNet & 0.21 & 0.60 & 2.14\\
            THOMAS~\cite{gilles2021thomas} & 0.20 & 0.64 & 2.16\\
            HDGT~\cite{jia2022hdgt} & 0.21 & 0.66 & 2.24 \\ 
            GNA & 0.29 & 0.71 & 2.45\\
            vilab & 0.29 & 0.71 & 2.47\\
            \hline 
    \end{tabular}
}
		\vspace{1mm}
\caption{The performance comparison on the test set leaderboard of the Argoverse 2 dataset. $K$ is the number of predicted trajectories for calculating the evaluation metrics.}
    \label{tab:argo2_test}
    \vspace{-3mm}
\end{table}

\subsection{Ablation Study}
We study the effectiveness of each component in our MTR/MTR++ frameworks. For efficiently conducting ablation experiments, we uniformly sampled 20\% frames (about 97k scenes) from the WOMD training set according to their default order\footnote{The detailed training data split can be found in our open-source codebase: \url{https://github.com/sshaoshuai/MTR}}, and we empirically find that it has similar distribution with the full training set. All models are evaluated with marginal motion prediction metric on the validation set of WOMD.

\begin{table}
    \centering\small
    \resizebox{0.48\textwidth}{!}{
        \begin{tabular}{cc|ccc}
            \hline
            \rowcolor{mygray}\tabincell{c}{Trajectory \\Generation} & \tabincell{c}{Iterative\\ Refinement} & minADE~$\downarrow$ & Miss Rate~$\downarrow$ & {\bf mAP}~$\uparrow$ \\
            \hline 
            \text{MLP} & & 0.6870 & 0.2103 & 0.2747	\\
            \text{Dense Goals} & & 1.0544 & 0.1936 & 0.2912 \\ 
            \text{Latent Embedding} & &  0.6564	& 0.1882 & 0.2826	\\ 
            \text{Intention Query} & & 0.6885 & 0.1723 & 0.3379		\\
            \text{Intention Query} & $\checkmark$ & \textbf{0.6557} & \textbf{0.1575} & \textbf{0.3539}	 \\
            \hline
        \end{tabular}
    }
    \vspace{1mm}
    \caption{Effects of different strategies for generating trajectories from encoded context features in the MTR framework.}
    \label{tab:motion_decoding}
    \vspace{-3mm}
\end{table}

\myparagraph{Effects of the learnable intention query.}
We investigate the effectiveness of different strategies for generating future trajectories based on encoded context features.  These strategies include the simple MLP head~\cite{jia2022hdgt,ngiam2021scene}, the goal-based head~\cite{gu2021densetnt}, the head with 6 latent anchor embeddings~\cite{varadarajan2021multipath++}, and the head with the learnable intention query. The first four rows of Table~\ref{tab:motion_decoding} illustrate the performance comparison of these strategies, where our proposed learnable intention query demonstrates significantly superior results. Specifically, our strategy achieves a much better mAP compared to the previous latent anchor embedding~\cite{varadarajan2021multipath++}  (\emph{i.e.}, +5.53\%) and dense-goal-based methods~\cite{zhao2020tnt,gu2021densetnt} (\emph{i.e.}, +4.67\%). This improvement can be attributed to our mode-specific querying strategy, where each intention query is associated with an explicit intention point, enabling more accurate and precise multimodal predictions.

\begin{table}
    \centering\small
    \resizebox{0.48\textwidth}{!}{
        \begin{tabular}{c|ccc}
            \hline
            \rowcolor{mygray}\tabincell{c}{Distribution of Intention Points}  & minADE~$\downarrow$ & Miss Rate~$\downarrow$ & {\bf mAP}~$\uparrow$ \\
            \hline 
            \text{uniform grids} & 0.7022 & 0.1952 & 0.3205	\\
            \text{k-means clustering} &  \textbf{0.6557} & \textbf{0.1575} & \textbf{0.3539}	 \\
            \hline
        \end{tabular}
    }
    \vspace{1mm}
    \caption{Effects of different strategies for generating intention points.}
    \label{tab:intention_points}
    \vspace{-3mm}
\end{table}

\begin{table}
    \centering\small
    \setlength\tabcolsep{10pt}
    \resizebox{0.48\textwidth}{!}{
        \begin{tabular}{cc|ccc}
        \hline
        \rowcolor{mygray}Attention & \#Polyline & minADE~$\downarrow$ & Miss Rate~$\downarrow$ & {\bf mAP}~$\uparrow$ \\
        \hline
        Global & 256 & 0.6701 & 0.1623 & 0.3450	\\
        Global & 512 & 0.6677 & 0.1610 & 0.3495	\\
        Global & 768 & OOM & OOM & OOM\\
        \cline{1-5}
        Local & 256 &  0.6692 & 0.1633 & 0.3522\\
        Local & 512 & 0.6685 & 0.1599 & 0.3515 \\
        Local & 768 & \textbf{0.6557} & 0.1575 & 0.3539 \\
        Local & 1024 & 0.6601 & \textbf{0.1555} & \textbf{0.3564} \\
        \hline
    \end{tabular}
    }
    \vspace{1mm}
    \caption{Effects of local self-attention in the transformer encoder of the MTR framework. ``\#polyline'' is the number of input map polylines used for context encoding. ``OOM'' indicates running out of memory.}
    \label{tab:localattn}
    \vspace{-5mm}
\end{table}

\myparagraph{Effects of the distribution of intention points.} 
As introduced in Sec.~\ref{sec:mtr_decoder}, we utilize the k-means clustering algorithm to generate 64 intention points, which serve as the foundation for our intention queries. In order to compare this approach with the straightforward uniform sampling strategy, we uniformly sample 8×8 = 64 intention points by considering the range of trajectory distribution for each category (see Fig.~\ref{fig:cluster}). The results presented in Table~\ref{tab:intention_points} indicate a significant drop in performance when replacing the k-means clustering algorithm with uniform sampling for generating intention points. This comparison highlights the superiority of our k-means clustering algorithm, as it produces a more accurate and comprehensive distribution of intention points. Consequently, it effectively captures the diverse future motion intentions of our interested agent with a small number of intention points.

\myparagraph{Effects of the iterative trajectory refinement.}
In Sec.~\ref{sec:mtr_encoder}, we introduce the utilization of stacked transformer decoder layers for iterative refinement of predicted trajectories by continually aggregating fine-grained features via dynamic map collection. As shown in the last two rows of Table~\ref{tab:motion_decoding}, this iterative refinement approach significantly reduces the miss rate metric by 1.48\% and improves the performance of mAP by +1.6\%. By continually aggregating trajectory-specific features from the context encoder with the proposed intention queries, the refinement process effectively improves the accuracy and quality of the predicted trajectories.

\begin{table*}[h]
    \centering\small
    \resizebox{0.98\textwidth}{!}{
        \begin{tabular}{lcc|cccc||ccc|ccc}
            \hline
            \rowcolor{mygray}  &  & & \multicolumn{4}{c||}{Performance with the Given Focal Agents} & \multicolumn{6}{c}{Efficiency with Increasing Numbers of Focal Agents} \\ 
            \cline{3-13}
            \rowcolor{mygray}  &  & & &&&& \multicolumn{3}{c|}{Inference Latency} & \multicolumn{3}{c}{Memory Cost } \\ 
            \rowcolor{mygray} \multirow{-3}{*}{Method} & \multirow{-3}{*}{\tabincell{c}{Symmetric \\ Context\\ Encoder}} & \multirow{-3}{*}{\tabincell{c}{Mutually-Guided\\Intention\\ Querying}} & \multirow{-2}{*}{minADE~$\downarrow$} & \multirow{-2}{*}{minFDE~$\downarrow$}& \multirow{-2}{*}{Miss Rate~$\downarrow$} & \multirow{-2}{*}{{\bf mAP}~$\uparrow$} & 8  & 16 & 32  & 8 & 16  & 32  \\ 
            \hline
            MTR & & & 0.6557 & 1.3362 & 0.1575 & 0.3539	  & 84ms & 123ms & 193ms & 5.2GB & 7.1GB & 15.6GB  \\ 
            \hline 
            MTR+& $\checkmark$  & & 0.6679 & 1.3486 & 0.1588 & 0.3505 & 67ms & 78ms & 98ms & 2.9GB & 3.2GB & 4.7GB	  \\
            MTR++ & $\checkmark$ & $\checkmark$ & \textbf{0.6490} &	\textbf{1.3163} &	\textbf{0.1559} & \textbf{0.3754} & 77ms & 90ms & 118ms & 3.1GB & 3.4GB & 5.2GB	 \\ 
            \hline
        \end{tabular}
    }
    \vspace{1mm}
    \caption{Effects of the symmetric scene context encoder and mutual-guided intention query in the MTR++ framework. The performance evaluation is conducted using a maximum of 8 given focal agents from the validation set of WOMD. To further assess the efficiency of different models as the number of focal agents increases, we set 8, 16, and 32 fake agents as the focal agents for calculating their inference efficiency and memory efficiency.}
    \label{tab:ab_mtr++}
    \vspace{-5mm}
\end{table*}

\myparagraph{Effects of local attention for context encoding.} 
Table~\ref{tab:localattn} demonstrates that the utilization of local self-attention in our context encoders leads to slightly superior performance compared to global attention when using the same number of map polylines as input. This finding confirms the significance of incorporating the input's local structure for more effective context encoding, and the inclusion of such prior knowledge through local attention positively impacts performance. Moreover, local attention proves to be more memory-efficient, allowing for performance improvements even when increasing the number of map polylines from 256 to 1,024. In contrast, global attention suffers from memory limitations due to its quadratic complexity.

\begin{figure}
    \centering
    \includegraphics[width=0.85\linewidth]{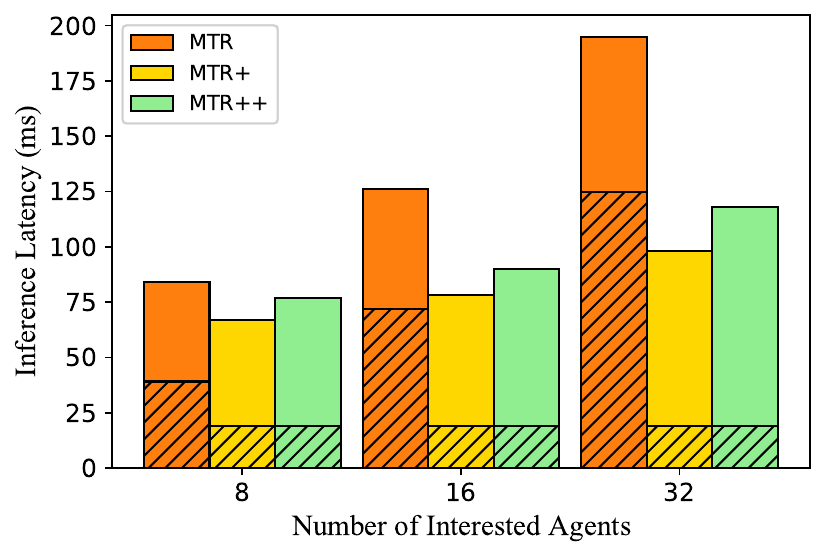}\\
		\vspace{-0.15in}
    \caption{The comparison of inference latency across different numbers of focal agents (\emph{i.e.}, interested agents) required to predict their future trajectories. The hatched area at the bottom of each pillar indicates the inference latency of the corresponding context encoder. MTR+ indicates the results obtained by only incorporating the symmetric context encoder into the MTR framework, while MTR++ indicates the results by further incorporating the mutually-guided intention querying strategy. 
}
    \label{fig:ab_inference_latency}
\end{figure}

\myparagraph{Effects of the symmetric scene context modeling module.}
In Sec.~\ref{sec:mtr++_encoder}, we present the symmetric scene context encoding module, which utilizes a shared context encoder for motion prediction of multiple interested agents in the same scene. Table~\ref{tab:ab_mtr++} demonstrates the effectiveness of incorporating our symmetric context encoder into the MTR framework (referred to as MTR+). With MTR+, we achieve comparable performance to the MTR framework while significantly reducing both inference latency and memory cost. Specifically, when the number of interested agents increases from 8 to 32, MTR requires individual scene context encoding for each agent, causing a substantial increase in inference latency and memory cost. In contrast, MTR+ utilizes the query-centric self-attention module to encode the entire scene with a shared symmetric context encoder, leading to a remarkable reduction in inference latency (from 193ms to 98ms for 32 interested agents) and memory cost (from 15.6GB to 4.7GB for 32 interested agents). Furthermore, we provide a breakdown analysis of inference latency in Fig.~\ref{fig:ab_inference_latency}, which demonstrates that as the number of interested agents increases, the latency of MTR's context encoder significantly rises, while the latency of MTR+'s context encoder remains constant due to the utilization of shared context features, since these shared context features enable the prediction of future trajectories for any number of agents within the scene.

\myparagraph{Effects of the mutually-guided intention querying strategy.}
Building upon our proposed symmetric scene context encoder for joint motion prediction, we introduce the mutually-guided intention querying strategy in Sec.\ref{sec:mtr++_decoder}. This strategy enables the interaction of future behaviors among multiple agents through the propagation of information among their intention queries. In Table\ref{tab:ab_mtr++}, we observe that the mutually-guided intention querying strategy significantly enhances the performance of MTR+ with a remarkable mAP improvement of $+2.49\%$. This improvement demonstrates the effectiveness of broadcasting the potential future behaviors of each agent to other agents via their intention queries, allowing MTR++ to predict more confident future behaviors by considering the overall development of the scene elements.

Furthermore, as each agent incorporates multiple intention queries (\emph{i.e.}, 64 in MTR frameworks), we investigate the interaction among these intention queries within each agent and across different agents. As presented in Table~\ref{tab:interaction_intention_query}, removing either type of interaction results in a significant decrease in performance by at least $-2.13\%$ in terms of mAP. Removing both types of interaction leads to a larger performance drop of $-2.70\%$ in terms of mAP. This analysis highlights the importance of the interaction among an agent's different intention queries, enabling the generation of more accurate multimodal future trajectories. Additionally, the interaction among intention queries across different agents empowers the model to predict informed and scene-compliant future trajectories for multiple agents, thereby yielding additional performance improvement.

\begin{table}
    \centering\small
    \resizebox{0.48\textwidth}{!}{
        \begin{tabular}{cc|ccc}
            \hline
             \rowcolor{mygray}\multicolumn{2}{c|}{Interaction of Intention Queries} & & &  \\
             \rowcolor{mygray}
             \tabincell{c}{Within Each Agent} & \tabincell{c}{Across Different Agents} &\multirow{-2}{*}{minADE~$\downarrow$} & \multirow{-2}{*}{Miss Rate~$\downarrow$} & \multirow{-2}{*}{{\bf mAP}~$\uparrow$}  \\
            \hline 
            & & 0.6708 &  0.1625 & 0.3484	\\ 
            $\checkmark$ &  & 0.6679 & 0.1588 & 0.3505 \\
            & $\checkmark$ & 0.6624 & 0.1566 & 0.3541	 \\
            $\checkmark$ & $\checkmark$ & \textbf{0.6490} & \textbf{0.1559} & \textbf{0.3754} \\
            \hline
        \end{tabular}
    }
    \vspace{1mm}
    \caption{Effects of the interaction of intention queries in the MTR++ framework. ``Within Each Agent'' and ``Across Different Agents'' indicates enabling the information interaction within the intention queries of each agent and across the intention queries of different agents, respectively.}
    \label{tab:interaction_intention_query}
    \vspace{-3mm}
\end{table}

\myparagraph{Effects of the query-centric self-attention.}
We introduce the query-centric self-attention module in Sec.~\ref{sec:mtr++_encoder}, which plays a vital role in modeling the relationship between tokens within their respective local coordinate systems. This module enables both symmetric scene context modeling and mutually-guided intention querying. In Table~\ref{tab:query_centric_self_attn}, we examine the effects of different positional encoding strategies in query-centric self-attention. The results in the first three rows indicate that query-centric relative positional encoding is crucial for achieving optimal performance. Removing this encoding or replacing it with global positional encoding significantly decreases performance by $-2.41\%$ and $-2.91\%$ in terms of mAP, respectively. This finding demonstrates the importance of modeling the relationship in the local coordinate system of each query token, as it benefits the simultaneous motion prediction for multiple agents by treating all tokens symmetrically. Additionally, comparing the last three rows of Table~\ref{tab:query_centric_self_attn}, we observe that adding positional embeddings to both the query/key tokens and value tokens yields the best performance.

\begin{table}
    \centering\small
    \resizebox{0.48\textwidth}{!}{
        \begin{tabular}{ccc|ccc}
            \hline
             \rowcolor{mygray}\multicolumn{3}{c|}{Positional Encoding in Self-Attention} & & & \\
             \rowcolor{mygray}
             \tabincell{c}{Strategy} & \tabincell{c}{With Query/Key} & \tabincell{c}{With Value} & \multirow{-2}{*}{minADE~$\downarrow$} & \multirow{-2}{*}{Miss Rate~$\downarrow$} & \multirow{-2}{*}{{\bf mAP}~$\uparrow$}  \\ 
            \hline 
             None & & & 0.6886 & 0.1614 & 0.3513 \\
             Global & $\checkmark$ & $\checkmark$  & 0.6913 &  0.1627 & 0.3463	\\
            Query-Centric & $\checkmark$ & $\checkmark$  & {\bf 0.6490} & {\bf 0.1559} & {\bf 0.3754} \\
            \hline 
           Query-Centric & $\checkmark$ & & 0.6523 & 0.1570 & 0.3658 \\
           Query-Centric & & $\checkmark$  & 0.6814 & 0.1603 & 0.3574\\
            \hline
        \end{tabular}
    }
    \vspace{1mm}
    \caption{Effects of the query-centric self-attention module in the MTR++ framework. ``With Query/Key'' and ``With Value'' indicates adding the position embedding to query/key tokens and value tokens, respectively.}
    \label{tab:query_centric_self_attn}
    \vspace{-3mm}
\end{table}

\begin{figure}
    \centering
    \includegraphics[width=0.99\linewidth]{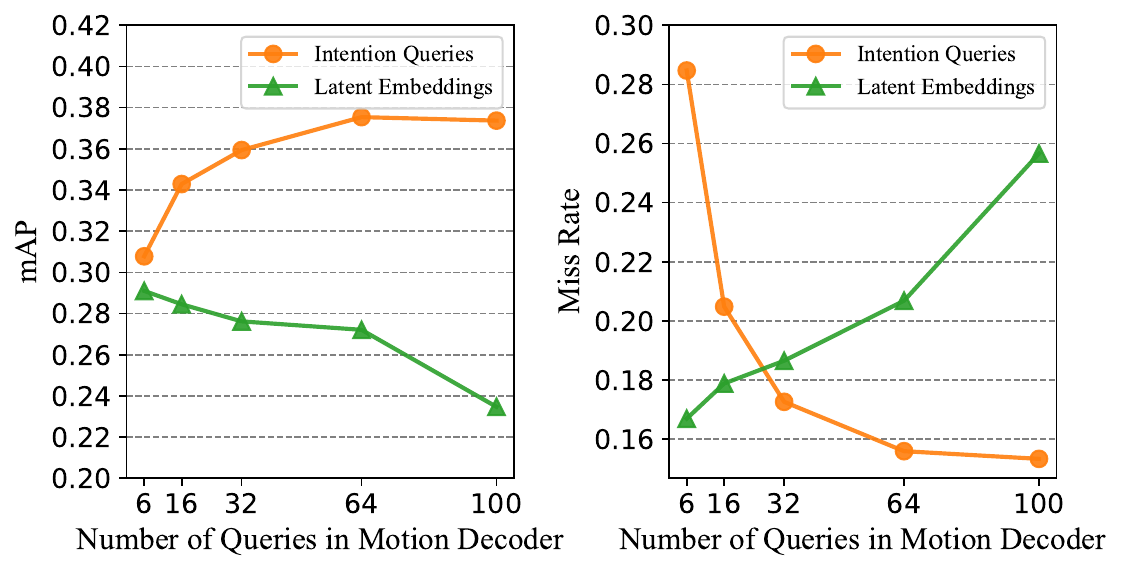}\\
		\vspace{-0.15in}
    \caption{The comparison of explicit intention queries and implicit latent embeddings in terms of different numbers of queries in the MTR++ framework. Two different colored curves demonstrate the performance of different strategies for generating future trajectories.}
    \label{fig:number_queries_two_strategies}
\end{figure}

\myparagraph{Effects of the number of intention queries.}
We conduct an ablation study to investigate the impact of the number of intention queries on the performance of the MTR++ framework. In Fig.~\ref{fig:number_queries_two_strategies}, we vary the number of intention queries by generating their intention points using the k-means clustering algorithm on the training dataset. The orange curves in Fig.~\ref{fig:number_queries_two_strategies} illustrate that the performance of the MTR++ framework improves significantly as the number of intention queries increases from 6 to 64. However, the performance saturates when the number of intention queries is further increased to 100. This ablation experiment highlights that incorporating 64 intention queries in the MTR frameworks already enables the coverage of diverse and wide-ranging future trajectories. This achievement is attributed to the design of learnable intention queries, which proves to be more efficient compared to previous goal-based strategies~\cite{zhao2020tnt,gu2021densetnt} that require a large number of goal candidates to achieve satisfactory performance.

\myparagraph{Discussion of explicit intention queries and implicit latent embeddings.}
In comparison to the latent anchor embeddings proposed in the state-of-the-art work MultiPath++\cite{varadarajan2021multipath++}, our proposed MTR frameworks establish a direct correspondence between intention queries and motion modes. This explicit mapping leads to faster convergence and improved performance. By referring to Fig.~\ref{fig:number_queries_two_strategies}, we can observe the following findings regarding the comparison between intention queries and latent embeddings with varying numbers of queries for the motion decoder: (1) Our strategy outperforms latent embeddings in terms of mAP and miss rate as the number of queries increases. This improvement is attributed to the fact that each intention query is specifically assigned to a particular motion mode, enabling a more stable training process. Conversely, in the case of latent embeddings, a ground truth trajectory can randomly associate with different anchor embeddings during training due to the lack of explicit correspondence. This randomness leads to training instability and decreased performance when increasing the number of anchor embeddings. (2) The explicit semantic interpretation of each intention query also contributes to its superior performance in terms of mAP. Intention queries are capable of predicting trajectories with more confident scores, thereby positively influencing the mAP metric. Overall, the establishment of explicit correspondence between intention queries and motion modes in our approach results in faster convergence, enhanced stability, and improved performance compared to previous latent embeddings.

\begin{table}[t]
    \centering\small
    \setlength\tabcolsep{10pt}
    \resizebox{0.48\textwidth}{!}{
        \begin{tabular}{c|ccc}
            \hline
            \rowcolor{mygray}\tabincell{c}{Dense Future Prediction}  & minADE~$\downarrow$ & Miss Rate~$\downarrow$ & {\bf mAP}~$\uparrow$ \\
            \hline 
             & 0.6662	& 0.1639 & 0.3606	\\
            \checkmark &  {\bf 0.6490} & {\bf 0.1559} & {\bf 0.3754}	 \\
            \hline
        \end{tabular}
    }
    \vspace{1mm}
    \caption{Effects of the dense future prediction module in the MTR++ framework.}
    \label{tab:dense_future_prediction}
    \vspace{-3mm}
\end{table}

\begin{table}[t]
    \centering\small
    \setlength\tabcolsep{10pt}
    \resizebox{0.48\textwidth}{!}{
        \begin{tabular}{c|ccc}
            \hline
            \rowcolor{mygray}\tabincell{c}{Number of Decoder Layers}  & minADE~$\downarrow$ & Miss Rate~$\downarrow$ & {\bf mAP}~$\uparrow$ \\
            \hline 
            1 & 0.6882	& 0.1689 & 0.3274 \\ 
            2 & 0.6665 & 0.1636 & 0.3561 \\ 
            3 &  0.6641	& 0.1653 & 0.3699 \\
            6 &  {\bf 0.6490} & {\bf 0.1559} & {\bf 0.3754}	 \\
            9 &  0.6512	& 0.1573 & 0.3728 \\ 
            \hline
        \end{tabular}
    }
    \vspace{1mm}
    \caption{Effects of the number of decoder layers in the MTR++ framework.}
    \label{tab:num_decoder_layer}
    \vspace{-3mm}
\end{table}

\begin{figure}
    \centering
    \includegraphics[width=0.92\linewidth]{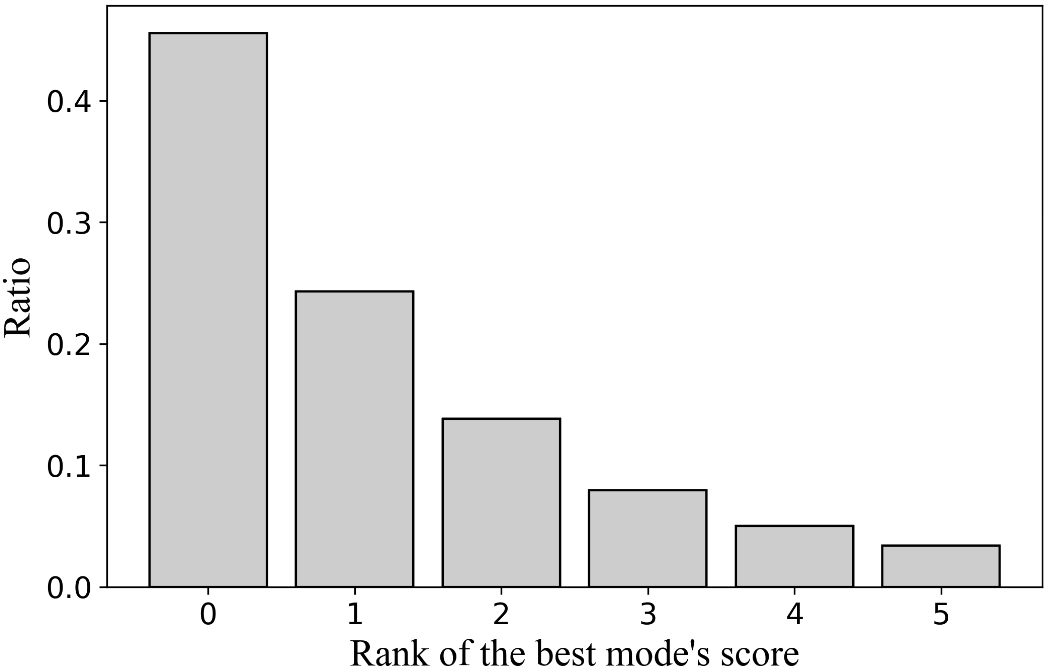}\\
    \vspace{-0.1in}
    \caption{The rank distribution of the best mode's score of our MTR++ framework on the Waymo Open Motion Dataset.}
    \label{fig:best_mode_distribution}
\end{figure}

\begin{figure*}
    \centering
    \includegraphics[width=0.98\linewidth]{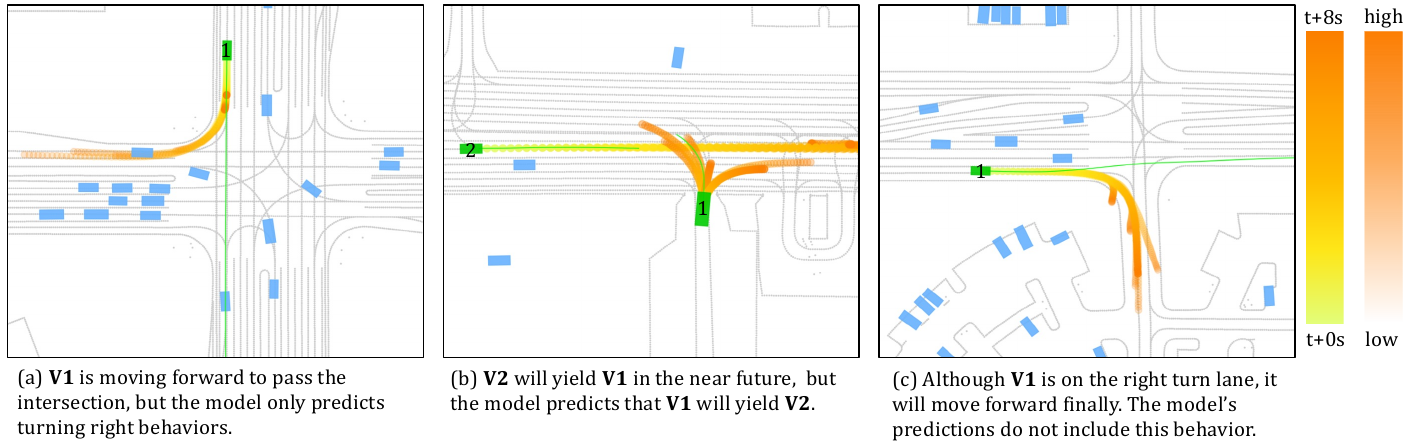}\\
    \vspace{-0.1in}
    \caption{Qualitative results of the failure cases of MTR frameworks on WOMD. The green curves indicate the ground-truth trajectories of our interested agents (green rectangle). We use gradient color to visualize the predicted trajectory waypoints at different future time steps, and trajectory confidence is visualized by setting
different transparent. Abbreviation: Vehicle (V).}
    \label{fig:failure_cases}
\end{figure*}

\myparagraph{Effects of dense future prediction.}
We investigate the impact of the dense future prediction module in Table~\ref{tab:dense_future_prediction}. By removing this module, we observe a significant decrease in the performance of the MTR++ framework, with a -1.48\% drop in mAP. We attribute this result to the beneficial effects of the dense future prediction module. It not only provides dense supervision for the context encoder, enabling it to learn more effective features for motion prediction of all agents in the scene, but also enhances the motion decoding process in the decoder network by incorporating agent features with their potential future trajectories, thereby enriching the contextual information for multimodal motion prediction.

\myparagraph{Effects of the number of decoder layers}. 
We investigate the number of transformer decoder layers in the MTR++ framework in Table~\ref{tab:num_decoder_layer}. We observe a consistent improvement in performance as we increase the number of decoder layers from 1 to 6. This improvement can be attributed to the stacked transformer decoder layers with the mutually-guided intention querying module, which facilitates the generation of more scene-compliant future trajectories through iterative trajectory refinement based on the predicted behaviors of other agents. However, increasing the number of decoder layers to 9 does not yield further improvement, suggesting a diminishing return. As a result, we adopt 6 decoder layers in our MTR++ framework to strike a balance between performance and efficiency.

\subsection{Discussion of Failure Cases and Future Challenges}
While MTR/MTR++ frameworks have demonstrated remarkable performance and achieved a state-of-the-art position, they continue to confront two primary challenges as below.

\myparagraph{Quality score estimation}: As evident in Table~\ref{tab:wod_test}, our model achieves a relatively lower average precision (i.e., below 0.5), while maintaining a favorable miss rate (approximately 0.13). This discrepancy arises due to the inadequacy of current quality score estimations in accurately reflecting the quality of each predicted trajectory, resulting in significant penalties imposed by the average precision metric. A compelling illustration of this issue is provided in Fig.~\ref{fig:best_mode_distribution}, where our analysis reveals that only 45.5\% of agents adhere to the predicted trajectory with the highest quality score. This underscores the limitations of existing quality score predictions, ultimately contributing to suboptimal average precision. 
Consequently, one important direction for future research involves investigating methods to refine the alignment between predicted scores and the quality of the corresponding trajectory. This would yield a more precise distribution of an agent's future behaviors, enhancing the predictive capabilities of our models.

\myparagraph{Diverse and accurate multimodal behaviors in rare scenarios:} Another significant challenge centers on generating diverse and accurate multimodal behaviors that faithfully capture the real intentions of the agent, particularly in rare scenarios. As illustrated in Fig.~\ref{fig:failure_cases}, our model, in certain scenarios, may generate homogenized future trajectories that fail to encompass the genuine intentions of the agent, resulting in imprecise multimodal predictions. Thus, a pivotal future research endeavor involves the development of methods that yield comprehensive multimodal behaviors, ensuring that multiple future trajectories not only exist but also represent distinct, meaningful intentions of the agent rather than merely variations along the same direction. 

These challenges and future research directions are instrumental in advancing the capabilities of motion prediction models, and addressing them will contribute to more accurate and comprehensive predictions, ultimately benefitting a wide range of applications.

\section{Conclusion}
In this paper, we have introduced the Motion TRansformer (MTR) frameworks as novel solutions for motion prediction in autonomous driving systems. The MTR frameworks employ a transformer encoder-decoder structure with learnable intention queries, effectively combining global intention localization and local movement refinement processes. This design enables the accurate determination of the agent's intent and adaptive refinement of predicted trajectories, resulting in efficient and precise prediction of multimodal future trajectories. Moreover, the proposed MTR++ framework enhances these capabilities by incorporating symmetric scene context modeling and mutually-guided intention querying modules, enabling the prediction of multimodal motion for multiple agents in a scene-compliant manner. Experimental results on the large-scale WOMD dataset demonstrate the state-of-the-art performance of the MTR frameworks on both marginal and joint motion prediction benchmarks. 

{\small
	\bibliographystyle{plain}
	\bibliography{egbib}
}

\vspace{5cm}
\begin{IEEEbiography}[{\includegraphics[width=1in,height=1.25in]{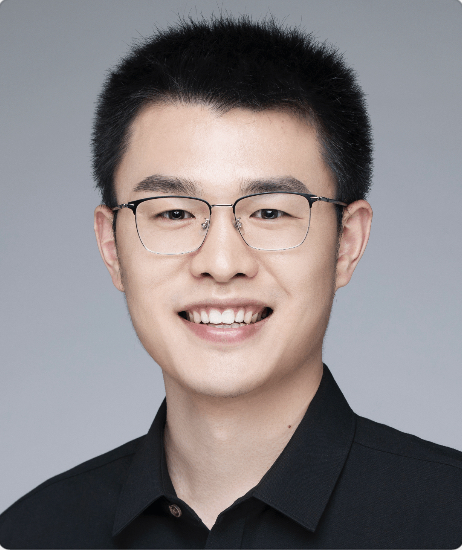}}]{Shaoshuai Shi} received his PhD degree in the Department of Electronic Engineering at The Chinese University of Hong Kong in 2021. He is currently a postdoctoral researcher at the Department of Computer Vision and Machine Learning of Max Planck Institute for Informatics. His research focuses on computer vision and machine learning, particularly in 3D scene understanding, object detection, motion prediction, knowledge transfer, as well as their applications in autonomous driving and robotics. 
\end{IEEEbiography}

\begin{IEEEbiography}[{\includegraphics[width=1in,height=1.25in]{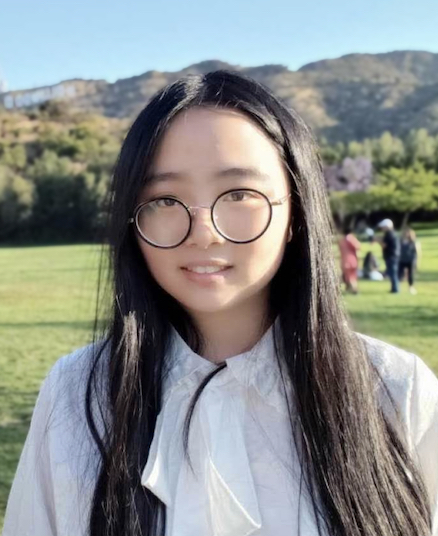}}]{Li Jiang} received her PhD degree in the Department of Computer Science and Engineering at The Chinese University of Hong Kong in 2021. She is currently an Assistant Professor at The Chinese University of Hong Kong, Shenzhen. Before that, She was a postdoctoral researcher at the Department of Computer Vision and Machine Learning of Max Planck Institute for Informatics. Her research interest includes computer vision and deep learning, particularly in 3D scene understanding, efficient representation learning, autonomous driving, and robotics.
\end{IEEEbiography}

\begin{IEEEbiography}[{\includegraphics[width=1in,height=1.25in]{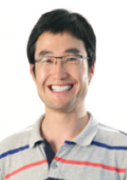}}]{Dengxin Dai} received the PhD degree in computer vision from ETH Zurich, Switzerland, in 2016. He was a Senior Researcher at MPI for Informatics and a Lecturer (external) at ETH Zurich. 
His research interests include autonomous driving, robust perception in adverse weather and illumination conditions, automotive sensors and computer vision under limited supervision. He has organized a CVPR Workshop series (’19, ’20) on Vision for All Seasons: Bad Weather and Nighttime, and has organized an ICCV’19 workshop on Autonomous Driving. He has been a program committee member of several major computer vision conferences and received multiple outstanding reviewer awards. He is also a guest editor for IJCV and the area chair for WACV’20 and CVPR’21.
\end{IEEEbiography}

\begin{IEEEbiography}[{\includegraphics[width=1in,height=1.25in]{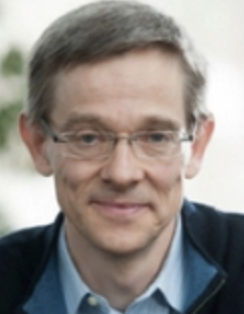}}]{Bernt~Schiele}(Fellow,~IEEE) received master's degrees both from the University of Karlsruhe, Germany, and ENSIMAG, Grenoble, France, in 1994. He received his PhD degree from INP Grenoble, France, in 1997. In 1994 he was a visiting researcher at Carnegie Mellon University, Pittsburgh, USA. From 1999 until 2004, he was an assistant professor with ETH Zurich and, from 2004 to 2010, he was a full professor of computer science with TU Darmstadt. In 2010, he was appointed a scientiﬁc member of the Max Planck Society and a director at the Max Planck Institute for Informatics. Since 2010, he has also been a professor with Saarland University. He is a fellow of IEEE, ACM, ELLIS and IAPR.
\end{IEEEbiography}

\end{document}